\documentclass{article} % For LaTeX2e
\usepackage{iclr2027_conference,times}

\usepackage{amsmath,amsfonts,bm}

\def\eqref#1{equation~\ref{#1}}
\def\1{\bm{1}}

\DeclareMathAlphabet{\mathsfit}{\encodingdefault}{\sfdefault}{m}{sl}
\SetMathAlphabet{\mathsfit}{bold}{\encodingdefault}{\sfdefault}{bx}{n}

\usepackage{hyperref}
\usepackage{url}
\usepackage[utf8]{inputenc} % allow utf-8 input
\usepackage[T1]{fontenc}    % use 8-bit T1 fonts
\usepackage{booktabs}       % professional-quality tables
\usepackage{amsfonts}       % blackboard math symbols
\usepackage{nicefrac}       % compact symbols for 1/2, etc.
\usepackage{microtype}      % microtypography
\usepackage{xcolor}         % colors

\usepackage{amsmath}
\usepackage{graphicx}
\usepackage{amssymb}
\usepackage{subcaption}
\usepackage{adjustbox}
\usepackage{makecell}
\usepackage{tabularx}
\usepackage{enumitem}
\usepackage{booktabs}
\usepackage{ulem}

\usepackage[table]{xcolor}
\usepackage{array}
\usepackage{caption}
\usepackage{bm}
\usepackage{longtable}

\colorlet{modelgray}{black!6}
\newcolumntype{N}{>{\centering\arraybackslash}X}
\newcolumntype{G}{>{\columncolor{modelgray}\centering\arraybackslash}X}

\title{Are We Really Benchmarking Forecasting Models? The Impact of Preprocessing on Time Series Performance}

\author{
  Guilherme Afonso Galindo Padilha \\
  LIVIA, Department of Software and IT Engineering \\
  École de technologie supérieure (ÉTS) \\
  Montreal, QC, Canada \\
  \texttt{guilherme-afonso.galindo-padilha.1@ens.etsmtl.ca} \\
  \And
  Paulo S. G. de Mattos Neto \\
  Center for Informatics (CIn) \\
  Universidade Federal de Pernambuco (UFPE) \\
  Recife, PE, Brazil \\
  \texttt{psgmn@cin.ufpe.br}
  \AND
  Rafael M. O. Cruz \\
  LIVIA, Department of Software and IT Engineering \\
  École de technologie supérieure (ÉTS) \\
  Montreal, QC, Canada \\
  \texttt{rafael.menelau-cruz@etsmtl.ca}
}

\iclrfinalcopy % Uncomment for camera-ready version, but NOT for submission.
\begin{document}

\maketitle

\begin{abstract}
While established literature underscores the pivotal role of preprocessing in forecasting accuracy, this stage remains largely overlooked in current research. Modern benchmarks typically resort to simple scaling, failing to account for critical transformations required to address nonstationarity, such as differencing. This omission creates a significant \textbf{structural preprocessing bias} that favors models with built-in data treatments while obscuring the true potential of simpler architectures. We study this effect through a preprocessing-aware benchmark that evaluates 11 forecasting models across 16 reversible preprocessing pipelines on 29,000 M4 time series. Our results identify preprocessing as a key driver of forecasting performance. Optimizing preprocessing per series yields gains of approximately 27\% to 87\% across all evaluated models, with architectures lacking internalized preprocessing experiencing the most substantial improvements. This allows simpler architectures to become highly competitive with complex, state-of-the-art models in modern forecasting benchmarks. All resources and experimental results from this benchmark are stored in a comprehensive metadataset to support future metalearning tasks. Codes~\footnote{\url{https://anonymous.4open.science/r/patm}} and data~\footnote{\url{https://kaggle.com/datasets/da12fb43d170e753a639014016fe33742660e87cfcb265b2e9a954186aa7fac5}} are available.
\end{abstract}

\section{Introduction}

Time series forecasting is a fundamental tool for decision-making in domains ranging from economics~\citep{economics} and finance~\citep{finances} to energy demand~\citep{energy} and healthcare~\citep{health, covid}. Because new forecasting methods are typically validated through benchmark comparisons, benchmarks play a central role in shaping research progress, determining which methods are perceived as state of the art, and guiding adoption in practice. Progress in the field therefore depends not only on developing new forecasting models, but also on the quality, transparency, and fairness of the benchmarks used to evaluate them.

However, forecasting performance is not determined by model architecture alone. It also depends on the preprocessing pipeline that transforms the raw time series before training and evaluation. This issue is especially important for nonstationary time series, which are common in real-world applications and may exhibit trends, evolving seasonality, or changing variance. Such properties alter the statistical conditions under which forecasting models are trained and tested. Many forecasting methods perform best when the input series is closer to stationary, while others incorporate mechanisms that partially compensate for nonstationarity internally~\citep{transformerstationarity, preprocessing:nonstationary, otexts}.

\begin{figure}
	\centering
	\includegraphics[width=0.75\textwidth]{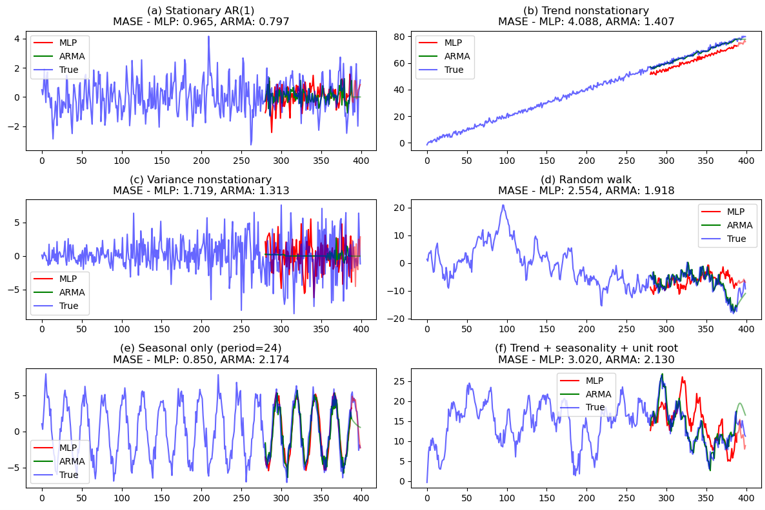}
	 \\ {\caption[Forecasts from an MLP and an ARMA without preprocessing. (a) Stationary AR(1): both models track the series well. (b) Trend nonstationarity: the rising level leads forecasts to lag the mean learned in training. (c) Variance nonstationarity: the mean is captured but larger fluctuations inflate absolute errors. (d) Random walk: persistent level changes cause drift and error accumulation. (e) Seasonality: without seasonal modeling, forecasts revert to the global mean and miss cyclic fluctuations. (f) Multi: multiple interacting forms of nonstationarity amplify drift and mismatch, causing the largest forecast degradation.]{Forecasts from an MLP and an ARMA without preprocessing. (a) Stationary AR(1): both models track the series well. (b) Trend nonstationarity: the rising level leads forecasts to lag the mean learned in training. (c) Variance nonstationarity: the mean is captured but larger fluctuations inflate absolute errors. (d) Random walk: persistent level changes cause drift and error accumulation. (e) Seasonality: without seasonal modeling, forecasts revert to the global mean and miss cyclic fluctuations. (f) Multi: multiple interacting forms of nonstationarity amplify drift and mismatch, causing the largest forecast degradation.}\label{fig:introduction}}
\end{figure}

Figure~\ref{fig:introduction} illustrates this dependency through forecasts from an MLP and an ARMA model under different data-generating regimes. Without a preprocessing pipeline, both models perform well in the stationary case, but their forecasts deteriorate as nonstationary components such as trend, changing variance, random-walk behavior, seasonality, or multiple interacting effects are introduced. Although this example is illustrative rather than a general proof, it highlights a central point of this work: preprocessing choices can materially change the effective difficulty of the forecasting task and, consequently, the apparent performance of different model families.

Despite the importance of preprocessing, many forecasting benchmarks either report preprocessing only partially or restrict it to minimal transformations such as scaling~\citep{benchmark:basicts, benchmark:gifteval, benchmark:monash, benchmark:tfb, benchmark:tsfm}. Transformations explicitly designed to address nonstationarity are rarely treated as systematic benchmark dimensions. As summarized in Table~\ref{tab:benchmarks}, commonly used benchmarks cover a broad range of models and datasets, but they generally do not evaluate forecasting algorithms under a controlled set of nonstationarity preprocessing pipelines.

\begin{table}[h!]
\caption{Benchmark characteristics. \textbf{X} = Yes; Blank = No. More in Appendix~\ref{appendix:benchmarks}.}
\centering
\begin{adjustbox}{width=\linewidth}
\begin{tabular}{lcccccccc}
\hline 
\textbf{Benchmark} &
\makecell{\textbf{Statistical}\\\textbf{Models}} &
\makecell{\textbf{Machine}\\\textbf{Learning}\\\textbf{Models}} &
\makecell{\textbf{Transformer-}\\\textbf{based}\\\textbf{Models}} &
\makecell{\textbf{Time Series}\\\textbf{Representation}} &
\makecell{\textbf{Scaling}\\\textbf{Preprocessing}} &
\makecell{\textbf{Nonstationarity}\\\textbf{Pipeline}} \\
\hline
Monash~\citep{benchmark:monash}     & X & X &   & X &   &   \\
BasicTS+~\citep{benchmark:basicts}   &   & X & X & X & X &   \\
TFB~\citep{benchmark:tfb}        & X & X & X & X & X &   \\
GIFT-EVAL~\citep{benchmark:gifteval}  & X & X & X & X & X &   \\
TSFM-Bench~\citep{benchmark:tsfm} & X & X & X & X & X &   \\
\hline
\textbf{Ours} & X & X & X & X & X & X \\
\hline
\end{tabular}
\end{adjustbox}
\label{tab:benchmarks}
\end{table}

This omission matters because preprocessing capacity is not distributed equally across forecasting methods. Some models include transformations that address nonstationarity as part of the model itself. For example, ARIMA incorporates differencing~\citep{timeseriesbook}, while recent neural approaches may include normalization mechanisms such as RevIN~\citep{revin}, as used in PatchTST~\citep{patchtst}. We refer to such methods as \textbf{I}nternalized \textbf{P}reprocessing Models (\textbf{IP-Models}), because part of the preprocessing burden is embedded within the model rather than applied externally.

When a benchmark applies only minimal external preprocessing, IP-Models may benefit from transformations unavailable to models without such built-in mechanisms. For example, recent studies have shown that RevIN can substantially influence the performance of state-of-the-art time series forecasting models~\citep{benchmark:probts, noiseorsignal}. As a result, benchmark outcomes may reflect not only architectural merit, but also unequal access to preprocessing capacity.

We refer to this phenomenon as \textbf{structural preprocessing bias}: a benchmark condition in which some models have access to preprocessing capabilities, either internally or externally, while others are evaluated under less favorable data conditions. This does not invalidate existing benchmarks. Rather, it limits the extent to which model-centric leaderboards can be interpreted as measuring forecasting architecture alone. \textbf{Without a controlled treatment of preprocessing, it remains unclear whether observed performance differences arise from the forecasting model, from the preprocessing implicitly available to that model, or from the interaction between the two.}

To address this gap, we propose a preprocessing-aware benchmarking framework for time series forecasting that decouples architectural merit from data transformation effects. The goal is to evaluate not only which models perform best under a fixed conventional pipeline, but also how different preprocessing strategies interact with different forecasting algorithms. In particular, we study how optimal and reversible preprocessing pipelines designed to handle nonstationarity affect the absolute performance and relative ranking of statistical, machine learning, and transformer-based forecasting methods.

The benchmark is conducted on a curated subset of the M4 competition dataset, comprising around 29,000 time series from multiple domains and frequencies~\citep{m4competition}. We evaluate 11 forecasting algorithms spanning statistical, machine learning, and transformer-based approaches, each combined with 16 reversible preprocessing pipelines.

The main contributions of this work are:

\begin{itemize}[noitemsep, topsep=0pt]
\item We propose a preprocessing-aware benchmark for time series forecasting that systematically combines diverse forecasting algorithms with diverse, reversible preprocessing pipelines for handling nonstationarity.
\item We analyze model sensitivity to preprocessing across statistical, machine learning, and transformer-based methods, showing that preprocessing can substantially alter both absolute performance and relative model rankings.
\item We release the complete collection of 5.1 million experimental results as a metadataset, enabling further analysis of the relationship between time series characteristics, preprocessing strategies, and forecasting performance.
\end{itemize}

\section{Time Series Preliminaries}

A time series is an ordered collection of observations indexed by time, $X=[x_1,x_2,…,x_n]$, usually recorded at regular intervals. The data may reflect continuous phenomena or discrete counts. For example, daily temperature is often treated as continuous, while the number of daily hotel visitors is discrete. Each value is time-dependent, meaning that its magnitude can be influenced by the time it was measured and by prior observations.

%\subsection{Time Series Characteristics}
%\label{sec:characteristics}
\noindent \textbf{2.1 Time Series Characteristics.}
A time series can be described by a set of features that summarize the temporal structure of the time series. Those descriptors give us a broader view to compare structures across datasets. Such representations are used in many fields, such as forecasting method selection~\citep{modelselection}, meta-learning~\citep{metalearning, m4metalearning}, classification and clustering~\citep{classification}, anomaly detection~\citep{anomaly}, etc. In this research, we describe key time series descriptors, including trend, seasonality, and stationarity~\citep{timeseriesbook}. The features used in this study are listed and described in Appendix~\ref{appendix:features}.

%\subsection{Time Series Preprocessing}

\noindent \textbf{2.2 Time Series Preprocessing.} Proper preprocessing is essential in time series forecasting because many traditional and machine learning estimators rely on specific statistical assumptions about their inputs~\citep{timeseriesbook}. When these assumptions are violated, parameter estimates, diagnostics, and predictive performance can degrade~\citep{otexts}. A primary objective is often achieving weak stationarity; thus, we apply transformations to remove or mitigate nonstationary components. For instance, as illustrated in Figure~\ref{fig:introduction} (d), a series exhibiting a unit root can be transformed via differencing to yield a series that is closer to stationarity (a), which can improve the performance of models that assume such properties.

Synthesizing standard practices in the field~\citep{otexts}, we categorize these transformations into (i) variance-stabilizing transforms, (ii) difference-based unit root removal, and (iii) scaling. Within the scope of this study, the selected transformations are invertible, allowing forecasts to be mapped back to the original scale. Appendix~\ref{appendix:preprocessing} contains a detailed description of the techniques used in this work.

%Proper preprocessing is essential in time series forecasting because many traditional and machine learning estimators rely on specific statistical assumptions about their inputs~\cite{timeseriesbook}. When these assumptions are violated, parameter estimates, diagnostics, and predictive performance can degrade~\cite{otexts}. A primary objective is often achieving weak stationarity; thus, we apply transformations to remove or mitigate nonstationary components. Here, $w_t$ denote the transformed series that is closer to a weak stationary series. Synthesizing standard practices in the field~\cite{otexts}, we categorize these transforms into (i) variance-stabilizing transforms, (ii) difference-based unit root removal, and (iii) scaling. Within the scope of this study, the selected transformations are invertible, allowing forecasts to be mapped back to the original scale.  Appendix~\ref{appendix:preprocessing} contains a description of the techniques used in this work.

%\subsubsection{Variance-Stabilizing Transforms}

\noindent \textbf{Internalized-Preprocessing Models.} We define an \textbf{IP-Model} as a composite architecture that maps an input sequence $s \subset X$ to a forecast $\hat{y}$ by embedding a preprocessing pipeline $p$ directly within the model's computational flow. Formally:
\begin{equation}
    \hat{y} = p^{-1}(F(p(s)))
\end{equation}
where $p$ represents a modular transformation (such as RevIN~\citep{revin}, differencing, or decomposition) designed to transform data, specifically nonstationarity, prior to the learning phase of the core forecasting algorithm $F$. This distinguishes IP-Models from standard architectures where $p$ is either absent or applied as a disjoint, external step. This differs from "structural features" (like patching or convolutions), which define the architecture's view of the data rather than its statistical state.

\section{Preprocessing-Aware Time Series Benchmark}

\begin{figure}
	\centering
	\includegraphics[width=0.8\textwidth]{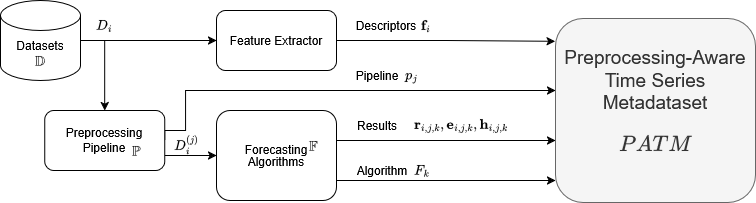}
	 \\ {\caption[Preprocessing-Aware Time Series Benchmark framework and Metadataset.]{Preprocessing-Aware Time Series Benchmark framework and Metadataset.}\label{fig:step1}}
\end{figure}

To analyze the impact of preprocessing on time series forecasting, we benchmark multiple forecasting algorithms under diverse preprocessing pipelines. All metadata and results are saved in a \textbf{P}reprocessing-\textbf{A}ware \textbf{T}ime Series \textbf{M}etadataset (PATM), shown in Figure~\ref{fig:step1}.

To ensure the benchmark captures performance across varied data regimes, we establish an experimental corpus $\mathbb{D} = \{D_1, D_2, \dots, D_\ell\}$. We contextualize each dataset $D_i$ by extracting a vector of descriptive features $\mathbf{f}_i = (f_i^1, f_i^2,\dots,f_i^z)$. These descriptors represent the structural properties of the series, allowing us to analyze how specific data characteristics interact with preprocessing choices to influence model success.

A core contribution of this benchmark is the formalization of the preprocessing search space. We define four critical categories of transformations: (1) $\mathbb{V}$: variance-stabilizing transformation techniques ($v_i\in\mathbb{V}$); (2) $\mathbb{U}$: unit-root removal techniques ($u_i\in\mathbb{U}$); (3) $\mathbb{S}$: seasonality removal techniques ($s_i\in\mathbb{S}$); (4) $\mathbb{Q}$: scaling techniques ($q_i\in\mathbb{Q}$).

% \begin{enumerate}
% \item $\mathbb{V}$: variance-stabilizing transformation techniques, where $v_i\in\mathbb{V}$;
% \item $\mathbb{U}$: unit-root removal techniques, where $u_i\in\mathbb{U}$;
% \item $\mathbb{S}$: seasonality removal techniques, where $s_i\in\mathbb{S}$;
% \item $\mathbb{Q}$: scaling techniques, where $q_i\in\mathbb{Q}$.
% \end{enumerate}

By considering the Cartesian product of these techniques, we define a pipeline space $\mathbb{P} = \mathbb{V} \times \mathbb{U} \times \mathbb{S} \times \mathbb{Q}$, with $|\mathbb{P}| := m$. Each dataset $D_i$ is processed through every pipeline $p_j\in\mathbb{P}$ to create a variant $D_i^{(j)} = p_j(D_i)$. This exhaustive design ensures full transparency and reproducibility while strictly preventing data leakage by fitting transformation parameters solely on training windows.

We evaluate a broad spectrum of forecasting algorithms $\mathbb{F} = \{F_1, F_2,\dots, F_n\}$, spanning statistical methods, classical machine learning, deep learning, and transformer-based architectures. For every preprocessed variant $D_i^{(j)}$, we execute a standardized training and inference protocol. Instead of accuracy-only leaderboards, we collect a multi-objective set of results for each experiment $(i, j, k)$: performance metrics $\mathbf{r}_{i,j,k}\in\mathbb{R}^r$, cost metrics $\mathbf{e}_{i,j,k}\in\mathbb{R}^e$, and configuration record $\mathbf{h}_{i,j,k}\in\mathbb{R}^h$ that contains model identity, and random seeds.

This benchmarking protocol enables a granular analysis of where preprocessing exerts the most significant impact—be it across dataset families, types of nonstationarity, or model architectures. We formalize the resulting metadataset generated by this extensive benchmark as the set of tuples: 

\begin{equation}\label{eq:patm}
PATM^\star = \left\{ (\mathbf{f_i}, p_j, F_k, \mathbf{r}_{i,j,k}, \mathbf{e}_{i,j,k}, \mathbf{h}_{i,j,k} ) \;\middle|\;
\begin{aligned}
& i \in \{1, \dots, \ell\}, \\
& j \in \{1, \dots, m\}, \\
& k \in \{1, \dots, n\}
\end{aligned}
\right\}.
\end{equation}

\section{Experimental Protocol}

\textbf{Hardware.} All experiments were run on a dedicated compute server. Detailed hardware and software specifications are reported in Appendix~\ref{appendix:hardware}.

\textbf{Datasets.} The evaluation focuses on univariate time series in order to isolate the effect of preprocessing. Multivariate settings introduce cross-channel interactions that can obscure the individual contribution of preprocessing to forecasting accuracy, as performance gains may arise from implicit information sharing rather than the transformations themselves. Moreover, recent work~\citep{patchtst} shows that modeling each channel independently can achieve competitive or even superior performance compared to approaches that explicitly model inter-variable dependencies. The M4 dataset~\citep{m4competition} provides an appropriate benchmark for this purpose, as it contains real-world time series from diverse application domains (e.g., Demographic, Macro, Industry, Finance)~\citep{benchmark:monash} and various temporal granularities. We utilize the quarterly, weekly, daily, and hourly subsets but exclude the yearly data. Yearly series are omitted because they lack a seasonal period, rendering seasonal differencing inapplicable and yielding no useful results for the analysis of seasonal transformations. The main characteristics of the dataset are shown in Table~\ref{tab:m4_characteristics}. Prior analyses~\citep{gratis, fformpp} have shown that the M4 dataset exhibits broad variability in trend behavior, seasonal patterns, autocorrelation profiles, and noise characteristics, which makes it suitable for assessing general forecasting performance across heterogeneous conditions. We adhere to the standard forecasting horizons defined in the M4 competition for each temporal granularity.

\begin{table}[t]
\caption{Main characteristics of the M4 datasets used in this study. Lag and horizon are reported in the number of time steps.}
\centering
\small
\resizebox{0.65\columnwidth}{!}{
\begin{tabular}{lcrrrrr}
\toprule
Dataset & Freq. & \# Series & Avg. Length & \# Obs. & Lag & Horizon \\
\midrule
M4 Quarterly & Quarterly & 24{,}000 & 100   & 2{,}406{,}108  & 4  & 8  \\
M4 Weekly    & Weekly    & 359      & 1{,}035 & 371{,}579     & 52 & 13 \\
M4 Daily     & Daily     & 4{,}227  & 2{,}371 & 10{,}023{,}836 & 7  & 14 \\
M4 Hourly    & Hourly    & 414      & 902   & 373{,}372      & 24 & 48 \\
\bottomrule
\end{tabular}
}
\label{tab:m4_characteristics}
\end{table}

\textbf{Pipelines.} We conduct a grid evaluation over the pipeline space $\mathbb{P}$. Operations are applied in a fixed sequence: variance-stabilization ($\mathbb{V}$), seasonal removal ($\mathbb{S}$), unit-root removal ($\mathbb{U}$), and scaling ($\mathbb{Q}$). We consider $\mathbb{V} \in \{\text{none}, \text{boxcox}\}$, $\mathbb{U} \in \{\text{none}, \text{diff}\}$, $\mathbb{S} \in \{\text{none}, \text{sdiff}\}$, and $\mathbb{Q} \in \{\text{none}, \text{z-score}\}$, resulting in 16 combinations. This sequence follows established frameworks~\citep{otexts}, ensuring variance is stabilized prior to differencing operations that can produce non-positive values. The setup prioritizes established, interpretable methods to isolate the systematic role of preprocessing without the interference of learned feature extractions.

\textbf{Forecasting Algorithms.} The selected model set includes \textbf{ETS}, \textbf{ARIMA}, and \textbf{Theta} as representatives of classical statistical forecasting.  \textbf{Linear Regression} and \textbf{XGBoost} serve as machine learning baselines. \textbf{MLP}, \textbf{N-BEATS}, and \textbf{TimesNet} represent deep learning architectures. \textbf{PatchTST} and \textbf{Informer} are also deep learning architectures, but are included as transformer-based models designed for high-capacity sequence modeling. Additionally, we evaluate \textbf{Chronos2-Synth}~\citep{chronos2}, a time series foundation model pre-trained exclusively on synthetic data. Foundation models represent a promising paradigm in forecasting with strong zero-shot generalization capabilities, making the analysis of their preprocessing sensitivity critical for future development. We specifically rely on a synthetically trained model to rigorously prevent data leakage. Since evaluating models trained on real-world corpora poses a significant risk that the M4 benchmark data was included in their pre-training set~\citep{rethinkingfm}, synthetic pre-training naturally avoids this overlap. This approach justifies the exclusion of other real-data foundation models and ensures that any observed performance variations reflect the model's true structural priors rather than memorization. Based on our proposed taxonomy, we classify \textbf{ARIMA}, \textbf{ETS}, \textbf{Theta}, \textbf{TimesNet}, \textbf{PatchTST}, and \textbf{Chronos2-Synth} as IP-Models. The total number of experiments is $11$ models $\times$ $16$ pipelines $\times$ approximately $29{,}000$ series $\approx 5.1 \times 10^6$ experiments. To strictly isolate the performance impact of preprocessing from model capacity, all hyperparameters and lag window lengths were adopted directly from established benchmark practices following the specifications documented by \citet{benchmark:monash}. Further details regarding this methodology and its trade-offs are provided in Appendix~\ref{appendix:hyperparameters}.

\textbf{Evaluation Metrics.} We record five performance metrics for every experiment: MSE, RMSE, MAE, sMAPE, and MASE. To maintain conciseness in our primary analysis and visualizations, we report the MASE as the primary metric. We prioritize MASE because it is strictly scale-independent, a necessity given that the M4 dataset aggregates series from disparate domains with different magnitudes.

\section{Results}
Table~\ref{tab:main_results} reports forecasting performance when a single preprocessing pipeline is selected and applied uniformly to the entire dataset. This setting reflects the standard design of many forecasting benchmarks, where a fixed and relatively simple preprocessing strategy, such as no transformation ($X$) or standard scaling ($Q$), is used for all models and all series~\footnote{Pipelines are composed of: \textbf{X} (Baseline/None), \textbf{V} (Boxcox transformation), \textbf{U} (First-order differencing), \textbf{S} (Seasonal differencing), and \textbf{Q} (Standard scaling).}. While this protocol simplifies evaluation, it can introduce severe mismatches between the chosen pipeline and the statistical properties of individual series. For instance, when combining $V$, $U$, and $S$ (such as in the VUS pipeline), error metrics often explode across several models. This occurs because the benchmark includes the full breadth of the M4 dataset, including non-seasonal series, without prior filtering. Applying seasonal and first-order differencing in such cases acts as a noise amplifier. When models produce erratic forecasts on these overly-differenced series, the inverse Box-Cox transformation exponentially amplifies these errors when returning the predictions to the original scale.%}{For instance, when combining $V$ and $S$, error metrics often explode across models. This occurs because the benchmark includes the full breadth of the M4 dataset, including non-seasonal series, without prior filtering; applying seasonal differencing in such cases acts as a noise amplifier, producing high-variance residuals that interact poorly with the non-linear stabilization of the Box-Cox transform.}

Under this evaluation protocol, IP-Models obtain the strongest overall results. In particular, when no preprocessing is applied, IP-Models achieve lower MASE than all non-IP-Models, and this advantage remains under standard scaling. This result supports the presence of \textbf{structural preprocessing bias}. Because IP-Models already include mechanisms that stabilize or adapt the input series, they are less dependent on the externally selected preprocessing pipeline. As a result, they remain competitive even when the benchmark uses a minimal or generic transformation. In contrast, non-IP-Models show stronger dependence on the chosen pipeline. For example, LR, XGBoost, and Informer improve substantially under differencing ($U$). This suggests that the performance of these models is disproportionately sensitive to whether the external preprocessing pipeline matches the series' statistical structure.

The best fixed-pipeline results further illustrate this distinction. The best IP-Model performance remains concentrated around simple or internally compatible pipelines, whereas the best non-IP-Model results require more specific preprocessing choices.  Therefore, when a benchmark evaluates all series under a single global pipeline, it tends to favor models that are resilient to preprocessing misspecification. This raises an important question: \textit{are IP-Models intrinsically better forecasters, or are they favored by benchmark protocols that use a single global preprocessing choice?}

\begin{table}[ht]
\caption{Forecasting performance (MASE) across preprocessing pipelines and models. Best per model in bold. Columns in grey are IP-Models. Pipelines are composed of: \textbf{X} (None), \textbf{V} (Boxcox transformation), \textbf{U} (First-order differencing), \textbf{S} (Seasonal differencing), and \textbf{Q} (Standard scaling).}

\centering
\small
\resizebox{0.8\columnwidth}{!}{
\begin{tabularx}{\textwidth}{l G G G N N N N G G N G}
\toprule
Pipeline & ARIMA & ETS & THETA & LR & XGBst & MLP & NBTS & TNet & PTST & Informer & Chronos \\
\midrule
X    & 2.35 & 2.25 & 2.22 & 2.87 & 2.75 & 31.88 & 4.65 & 2.65 & 2.77 & 106.75 & 2.29 \\
V    & 3.18 & 2.38 & 2.36 & 3.75 & 2.61 & 4.27 & 4.78 & \textbf{2.45} & \textbf{2.48} & 46.12 & 2.56 \\
U    & 2.23 & 2.40 & 2.60 & \textbf{2.27} & 2.34 & 3.10 & 5.91 & 3.08 & 3.16 & 3.12 & \textbf{2.09} \\
S    & 2.38 & 3.19 & 3.19 & 2.71 & 2.49 & 4.06 & 5.78 & 3.10 & 3.42 & 4.45 & 2.46 \\
Q    & 2.30 & \textbf{2.18} & \textbf{2.22} & 2.87 & 3.71 & 4.86 & 4.86 & 2.54 & 2.66 & 5.51 & 2.29 \\
VU   & 2.23 & 2.45 & 2.84 & 2.45 & 2.18 & 2.87 & $1.48e^{24}$ & 3.35 & 3.47 & 3.21 & 2.17 \\
VS   & 2.81 & 3.85 & 3.83 & $2.35e^{4}$ & 2.34 & 2.81 & 2.72 & 3.20 & 3.22 & 2.74 & 2.66 \\
VQ   & 2.28 & 2.38 & 2.38 & 3.75 & 3.78 & 5.47 & 2.66 & 2.54 & 2.62 & 5.31 & 2.35 \\
US   & 2.44 & 3.81 & 4.47 & 3.13 & 2.86 & 3.31 & 9.82 & 6.60 & 6.53 & 6.45 & 3.28 \\
UQ   & \textbf{2.20} & 2.34 & 2.60 & 2.27 & \textbf{2.16} & \textbf{2.21} & 5.80 & 3.08 & 3.04 & \textbf{2.25} & \textbf{2.09} \\
SQ   & 2.39 & 3.13 & 3.19 & 2.71 & 2.36 & 2.58 & \textbf{2.53} & 3.03 & 3.02 & 2.49 & 2.46 \\
VUS  & 2.57 & 27.02 & $2.99e^{4}$ & $1.38e^{37}$ & 2.95 & 3.80 & $4.97e^{35}$ & $7.19e^{4}$ & $2.42e^{5}$ & 3.70 & 5.27 \\
VUQ  & 2.33 & 2.45 & 2.85 & 2.45 & 2.36 & 2.55 & $1.38e^{24}$ & 3.22 & 3.54 & 2.45 & 2.17 \\
VSQ  & 2.91 & 3.59 & 3.83 & $2.35e^{4}$ & 2.54 & 2.74 & 3.73 & 3.24 & 3.14 & 2.75 & 2.66 \\
USQ  & 2.72 & 3.65 & 4.47 & 3.13 & 2.97 & 3.03 & 3.46 & 6.34 & 6.47 & 3.17 & 3.28 \\
VUSQ & 3.22 & 27.01 & $3.01e^{4}$ & $1.38e^{37}$ & 3.27 & 3.13 & $5.19e^{35}$ & $8.96e^{6}$ & $1.56e^{5}$ & 3.85 & 5.28 \\
\bottomrule
\end{tabularx}
}
\label{tab:main_results}
\end{table}

\begin{figure}[h] \centering \includegraphics[width=0.75\textwidth]{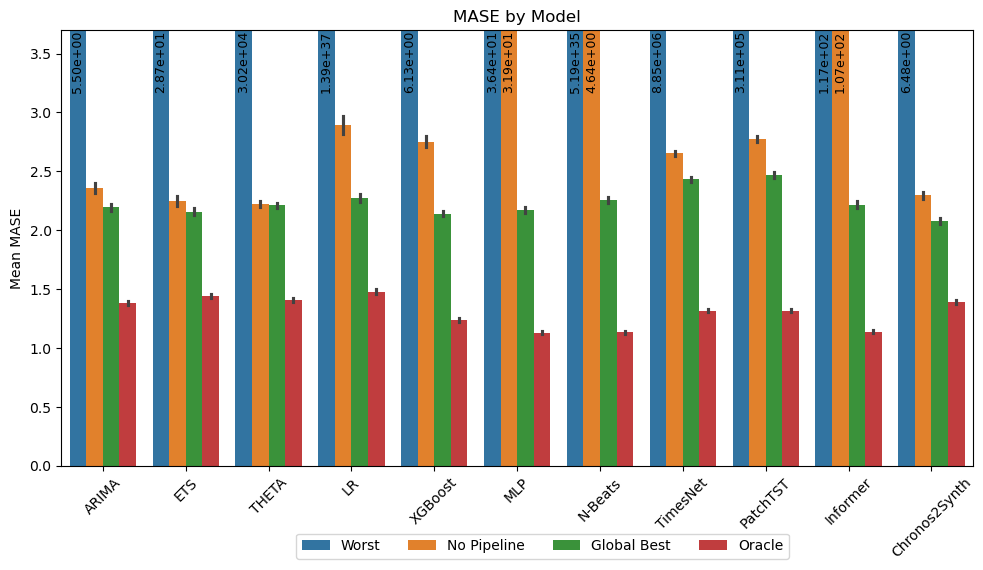} \caption{Barplot MASE of best and oracle pipelines. Lower loss values mean better results.} \label{fig:oracle_mase} \end{figure}

\textbf{Per-series preprocessing significantly impacts model performance.}
Figure~\ref{fig:oracle_mase} compares each model under four preprocessing settings: \emph{No Pipeline}, the single \emph{Best Global} pipeline, the \emph{Worst} global pipeline, and an \emph{Oracle} configuration in which the best preprocessing pipeline is selected separately for each time series. The comparison shows two distinct behaviors. First, IP-Models exhibit only small differences between the \textit{No Pipeline} and \textit{Global Best} settings, reinforcing that these models are relatively robust to preprocessing misspecification and can perform well even when little or no external preprocessing is applied. In contrast, most non-IP-Models already improve when moving from no preprocessing to the best global pipeline, indicating that evaluating them on raw data places them at a disadvantage.

The \textit{Oracle} setting further amplifies this difference. Since the oracle selects the best pipeline per series, all models improve by definition. However, the magnitude of improvement remains much larger for non-IP-Models. While IP-Models show substantial oracle gains against the no pipeline baseline (ranging from approximately 35\% to 55\%), the improvement is even more dramatic for non-IP-Models. Specifically, four out of the five non-IP-Models improve by more than 50\%, with models like MLP and Informer improving by over 75\%. In some cases, this allows non-IP-Models to close the performance gap with IP-Models. \textbf{These results suggest that the strong relative standing of IP-Models under fixed benchmark pipelines is partly explained by their robustness to generic preprocessing, whereas non-IP-Models benefit more strongly when preprocessing is adapted to the statistical properties of each series.} Therefore, neglecting per-series preprocessing can create an unfair comparison by systematically underestimating models that rely more heavily on external data transformations. Additional results are provided in Appendix~\ref{appendix:nopipelinexoracle}.
%}{The \textit{Oracle} setting further amplifies this difference. Since the oracle selects the best pipeline per series, all models improve by definition. However, the magnitude of improvement is much larger for non-IP-Models. While the largest oracle gain (against no pipeline) among IP-Models is observed for ETS, at approximately 25\%, four out of the five non-IP-Models improve by more than 35\%. In some cases, this allows non-IP-Models to close the performance gap with IP-Models. \textbf{These results suggest that the strong relative standing of IP-Models under fixed benchmark pipelines is partly explained by their robustness to generic preprocessing, whereas non-IP-Models benefit more strongly when preprocessing is adapted to the statistical properties of each series.} Therefore, neglecting per-series preprocessing can create an unfair comparison by systematically underestimating models that rely more heavily on external data transformations. Additional results are provided in Appendix~\ref{appendix:nopipelinexoracle}.}

%\textcolor{red}{std looks terrible, so I added CI. mention it in the text and give an interpretation.}

% \begin{figure}[h]
%     \centering
%     \includegraphics[width=0.95\textwidth]{figures/mase_hourly.png}
%     \caption{Barplot MASE of best and oracle pipelines. Lower loss values means better results.}\label{fig:oracle_mase}
% \end{figure}

\begin{figure}[ht]
\centering

\begin{minipage}[t]{0.40\textwidth}
    \vspace{0pt}
    \centering
    \includegraphics[width=\linewidth]{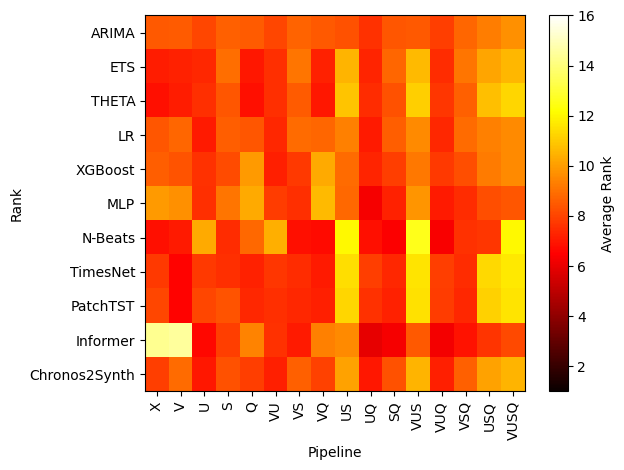}
    \caption{Intra-model average rank of each preprocessing pipeline. Each model is ranked only against itself across pipelines; lower ranks indicate better preprocessing choices for that model.}
    \label{fig:rank_heatmap}
\end{minipage}
\hfill
\begin{minipage}[t]{0.58\textwidth}
    \vspace{0pt}
    \captionof{table}{Inter-model mean rank ($\pm$ std) under selected preprocessing configurations. Lower ranks indicate better relative performance. Grey rows indicate IP-Models. Bold values indicate the three best-ranked models within each column, including ties.}
    \label{tab:leaderboard_results2}
    \centering
    \small
    \resizebox{\linewidth}{!}{
\begin{tabular}{l c c c c c c}
\toprule
& Oracle & X & V & U & S & Q \\
\midrule
\rowcolor{gray!12}
ARIMA 
& 5.94 {\tiny $\pm$3.28} 
& 4.99 {\tiny $\pm$2.54} 
& \textbf{5.15} {\tiny $\pm$2.80} 
& 5.30 {\tiny $\pm$3.03} 
& \textbf{4.99} {\tiny $\pm$3.15} 
& 4.93 {\tiny $\pm$2.65}  \\

\rowcolor{gray!12}
ETS 
& 6.70 {\tiny $\pm$3.00} 
& \textbf{4.61} {\tiny $\pm$2.66} 
& \textbf{4.73} {\tiny $\pm$2.66} 
& \textbf{5.25} {\tiny $\pm$2.76} 
& 6.18 {\tiny $\pm$2.89} 
& \textbf{4.51} {\tiny $\pm$2.81}  \\

\rowcolor{gray!12}
THETA 
& 6.76 {\tiny $\pm$3.08} 
& \textbf{4.73} {\tiny $\pm$2.52} 
& \textbf{4.97} {\tiny $\pm$2.53} 
& 5.79 {\tiny $\pm$2.98} 
& 6.51 {\tiny $\pm$3.00} 
& \textbf{4.74} {\tiny $\pm$3.02}  \\

LR 
& 6.87 {\tiny $\pm$3.34} 
& 5.54 {\tiny $\pm$3.15} 
& 5.66 {\tiny $\pm$3.15} 
& \textbf{5.23} {\tiny $\pm$3.01} 
& 5.38 {\tiny $\pm$3.17} 
& 5.61 {\tiny $\pm$3.38}  \\

XGBoost 
& 5.48 {\tiny $\pm$2.89} 
& 5.47 {\tiny $\pm$2.99} 
& 5.47 {\tiny $\pm$2.72} 
& 5.28 {\tiny $\pm$2.61} 
& \textbf{5.06} {\tiny $\pm$2.68} 
& 6.38 {\tiny $\pm$2.49}  \\

MLP 
& \textbf{4.88} {\tiny $\pm$2.72} 
& 7.34 {\tiny $\pm$2.97} 
& 6.92 {\tiny $\pm$3.46} 
& 6.17 {\tiny $\pm$3.00} 
& 6.65 {\tiny $\pm$3.47} 
& 7.58 {\tiny $\pm$2.89}  \\

N-BEATS 
& \textbf{5.39} {\tiny $\pm$3.09} 
& 5.63 {\tiny $\pm$2.96} 
& 5.93 {\tiny $\pm$3.06} 
& 8.53 {\tiny $\pm$3.40} 
& 6.53 {\tiny $\pm$3.07} 
& 7.65 {\tiny $\pm$3.41}  \\

\rowcolor{gray!12}
TimesNet 
& 6.25 {\tiny $\pm$3.27} 
& 6.05 {\tiny $\pm$2.64} 
& 5.59 {\tiny $\pm$2.53} 
& 6.68 {\tiny $\pm$3.26} 
& 6.14 {\tiny $\pm$3.10} 
& 5.74 {\tiny $\pm$2.59}  \\

\rowcolor{gray!12}
PatchTST 
& 6.15 {\tiny $\pm$3.23} 
& 6.40 {\tiny $\pm$2.78} 
& 5.71 {\tiny $\pm$2.45} 
& 6.87 {\tiny $\pm$3.27} 
& 6.87 {\tiny $\pm$3.37} 
& 5.82 {\tiny $\pm$2.47}  \\

Informer 
& \textbf{5.23} {\tiny $\pm$2.92} 
& 10.35 {\tiny $\pm$2.01} 
& 10.40 {\tiny $\pm$1.98} 
& 6.17 {\tiny $\pm$2.96} 
& 6.61 {\tiny $\pm$3.21} 
& 8.18 {\tiny $\pm$3.16}  \\

\rowcolor{gray!12}
Chronos2Synth
& 6.34 {\tiny $\pm$3.18} 
& \textbf{4.90} {\tiny $\pm$2.78} 
& 5.37 {\tiny $\pm$2.97} 
& \textbf{4.68} {\tiny $\pm$2.44} 
& \textbf{5.01} {\tiny $\pm$2.68} 
& \textbf{4.85} {\tiny $\pm$2.99}  \\
\bottomrule
\end{tabular}
    }
\end{minipage}

\end{figure}

\textbf{Preprocessing changes model rankings.} Figure~\ref{fig:rank_heatmap} analyzes preprocessing preferences through an intra-model ranking, where each model is compared only against its own performance under different pipelines. This view reveals two distinct patterns. IP-Models generally obtain their best ranks under simple pipelines such as $X$, $V$, $Q$, and $VQ$, while heavily stacked transformations such as $VUS$, $USQ$, and $VUSQ$ tend to perform poorly. This reinforces that these models are relatively robust to minimal preprocessing and do not usually require aggressive external transformations.

% \begin{figure}[h]
% \centering
% \includegraphics[width=0.50\textwidth]{figures/averagerank.png}
% \caption{Intra-model average rank of each preprocessing pipeline. Each model is ranked only against itself across pipelines; lower ranks indicate better preprocessing choices for that model.}
% \label{fig:rank_heatmap}
% \end{figure}

Non-IP-Models show a more heterogeneous profile. Their preferred pipelines vary substantially across architectures. For example, $U$ is beneficial for LR but extremely harmful for N-BEATS. Nevertheless, most non-IP-Models benefit from pipelines involving first-order differencing ($U$), suggesting that external stationarity-inducing transformations play a larger role for models without internalized preprocessing mechanisms. Thus, the heatmap shows that preprocessing is not only important in magnitude, but also model-dependent in direction. A more detailed, per-model view is shown in Appendix~\ref{appendix:ranking}.

Table~\ref{tab:leaderboard_results2} shifts from this intra-model view to an inter-model ranking, where models are compared against each other under the same preprocessing configuration. Under the commonly used $Q$ pipeline~\citep{benchmark:tfb, benchmark:monash, benchmark:basicts, benchmark:gifteval, benchmark:tsfm}, the best ranks are dominated by IP-Models. However, under the oracle setting, the relative positioning shifts favorably for several non-IP-Models. Although this comparison should not be interpreted as a fully optimized leaderboard, since no model-specific hyperparameter search is performed, it shows that relative model standings can change substantially once preprocessing is selected per series. This supports our central claim: fixed preprocessing protocols can favor models that are robust to generic pipelines, while underestimating models whose performance depends more strongly on appropriate external transformations.

\textbf{Preprocessing Preferences Depend on Model and Series Properties.} Table~\ref{tab:recommendations} reports the single preprocessing pipeline that achieved the best overall MASE for each model across different data characteristics. This analysis demonstrates that preprocessing effectiveness is not only model-dependent but also strongly conditioned by the statistical properties of the input series. For a more granular visualization showing the frequency of pipelines appearing in the Top-3, see Appendix~\ref{appendix:heatmap}.

\begin{table}[h]
\caption{Best performing preprocessing pipelines (MASE) for each model under different statistical properties. IP-Models are highlighted in gray.}
\label{tab:recommendations}
\centering
\small
\resizebox{0.8\columnwidth}{!}{
\begin{tabular}{l c c c c c c c c c c}
\toprule
& \multicolumn{2}{c}{\textbf{Trend}} & \multicolumn{2}{c}{\textbf{Seasonality}} & \multicolumn{2}{c}{\textbf{Stationarity}} & \multicolumn{2}{c}{\textbf{Shifting}} & \multicolumn{2}{c}{\textbf{Entropy}} \\
\cmidrule(lr){2-3} \cmidrule(lr){4-5} \cmidrule(lr){6-7} \cmidrule(lr){8-9} \cmidrule(lr){10-11}
\textbf{Model} & High & Low & High & Low & Stat. & Non-Stat. & High & Low & High & Low \\
\midrule
\rowcolor{gray!12}
ARIMA     & VUQ & VS & VUQ & VUQ & VUQ & VUQ & VUQ & VUQ & VQ & VUQ \\
\rowcolor{gray!12}
ETS       & X & Q & Q & X & V & VQ & Q & VQ & V & V \\
\rowcolor{gray!12}
THETA     & X & X & X & \textbf{X} & \textbf{X} & \textbf{X} & \textbf{X} & \textbf{VQ} & Q & Q \\
LR        & VU & V & VU & VU & UQ & VU & VU & Q & V & VUQ \\
XGBoost   & VUQ & VQ & VSQ & VUQ & SQ & VUQ & VUQ & VUQ & VQ & VUQ \\
MLP       & VUQ & VSQ & VSQ & UQ & VUSQ & VUQ & VUQ & Q & VQ & USQ \\
N-BEATS   & VUQ & SQ & SQ & UQ & SQ & VUQ & SQ & VUQ & Q & SQ \\
\rowcolor{gray!12}
TimesNet  & VQ & \textbf{Q} & \textbf{Q} & VQ & Q & VQ & VQ & VQ & VQ & Q \\
\rowcolor{gray!12}
PatchTST  & VQ & VQ & SQ & VQ & VQ & VQ & VQ & Q & \textbf{VQ} & V \\
Informer  & \textbf{VUQ} & SQ & USQ & VU & UQ & VUQ & VUQ & VQ & VQ & \textbf{VUQ} \\
\rowcolor{gray!12}
Chronos2Synth & VUQ & VQ & Q & VUQ & Q & VUQ & Q & X & Q & VU \\
\bottomrule
\end{tabular}
}
\end{table}

The clearest patterns emerge from the trend and non-stationarity splits. When trend strength is low or the series is stationary, most models perform best with simple pipelines such as $X$, $V$, $Q$, and $VQ$. When trend strength is high or the data is highly non-stationary, however, non-IP-Models shift toward pipelines involving first-order or seasonal differencing. For example, XGBoost, MLP, N-BEATS, and Informer overwhelmingly favor variants like $VUQ$, $UQ$, or $USQ$ under high trend and non-stationarity. This behavior aligns with the classical role of differencing in removing trend-induced drifts. In contrast, IP-Models like THETA, TimesNet, and PatchTST preserve much simpler preprocessing preferences across these challenging subsets, indicating that their internal mechanisms successfully reduce their dependence on heavy external transformations. A notable exception is ARIMA, which strongly prefers the VUQ pipeline, likely because external variance stabilization and scaling help the algorithm more stably identify and fit the appropriate internal integration order.

The stationarity and shifting splits produce results that further validate our core argument. Because standard stationarity tests do not distinguish whether the difficulty arises primarily from trend, seasonality, or variance changes, the preferred pipelines for the non-stationary subset are typically comprehensive transformations like $VUQ$. These robust pipelines address multiple forms of degradation simultaneously. This analysis shows that structural preprocessing bias emerges precisely because different forms of non-stationarity affect model families differently, and fixing a single global pipeline will inevitably penalize models that lack internal stabilization mechanisms.

\begin{figure}[h]
\centering
\includegraphics[width=0.65\textwidth]{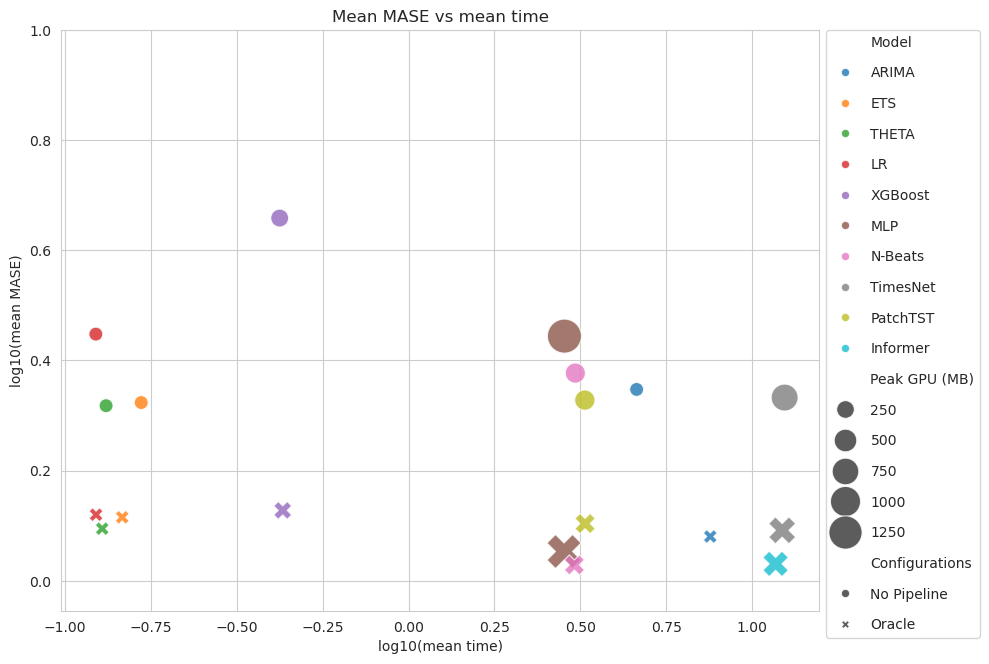}
\caption{Mean MASE vs. mean training time for local models. Marker size indicates peak GPU memory (baselined at 250MB for CPU-only models).}
\label{fig:mase_time}
\end{figure}

\textbf{Cost accounting reveals efficiency trade-offs that preprocessing can optimize.} Figure~\ref{fig:mase_time} exposes the trade-off between predictive accuracy, training time, and peak GPU memory usage. To calculate the computational impact fairly and independently of dataset benchmark structure, we evaluate models by training them locally on each individual series. Deep learning models often exhibit substantially longer training times and higher memory footprints. However, when optimal preprocessing is applied, classical models like LR and THETA, as well as simpler machine learning models, achieve highly competitive performance with a fraction of the computational cost. Importantly, applying these preprocessing pipelines provides a significant performance boost with virtually no practical increase in training time. For a broader perspective on cost when machine learning models are trained globally across multiple series simultaneously, refer to Appendix~\ref{appendix:cost}. Note that Chronos2Synth is excluded from this figure because it is a zero-shot model and requires no training.

\section{Conclusion}

This paper introduced a preprocessing-aware benchmark for time series forecasting, showing that model performance and rankings are strongly shaped by the preprocessing pipeline used during evaluation. Across statistical, machine learning, and transformer-based models, we found that fixed global pipelines favor models with internalized preprocessing mechanisms, while models without such mechanisms benefit substantially more when preprocessing is selected per series. We refer to this effect as \textbf{structural preprocessing bias}: an evaluation artifact in which benchmark outcomes reflect unequal access to preprocessing capacity in addition to architectural merit. Our results show that preprocessing preferences depend on both model class and series properties, with trend and seasonality inducing clear shifts toward differencing-based pipelines for non-IP-Models. These findings suggest that forecasting benchmarks should treat preprocessing as an explicit evaluation dimension rather than a secondary implementation detail. To support further work, we release the PATM metadataset as a resource for studying preprocessing sensitivity and developing adaptive preprocessing selection methods.

\subsection*{AI use statement}

(This section is \textbf{required} and does not count toward the page limit.)

In this work, we used generative AI tools to assist in editing the research paper to improve readability, format and create tables, and edit software code by polishing, commenting, and refactoring it. We have not used generative AI tools for generating synthetic datasets, developing theoretical models or conceptual frameworks, formulating mathematical claims, proving mathematical claims, proposing or refining hypotheses, designing or providing feedback on research methodology or experiments, interpreting results, or any other tasks with required disclosure. We have reviewed all AI-assisted work. AI-generated text and tables were manually verified for accuracy and consistency with our findings. LLM-generated and refactored code was thoroughly reviewed and tested for correctness by the authors. We take responsibility for the final content of this work, including text, claims, and artifacts produced with the aid of generative AI.

\subsection*{Ethics statement}

(This section is \textbf{recommended} and does not count toward the page limit.)

This research does not involve human subjects, nor does it rely on sensitive or proprietary data. We rely exclusively on publicly available datasets that have been anonymized and widely adopted in the forecasting literature. We believe there are no significant ethical concerns, potential conflicts of interest, privacy issues, or risks of harm associated with this work and its applications.

\subsection*{Reproducibility statement}

(This section is \textbf{recommended} and does not count toward the page limit.)

To ensure the full reproducibility of our benchmark and findings, we provide comprehensive details and open-source materials. We outline our experimental protocol, including dataset characteristics, model selections, and evaluation procedures, in Section 4. Detailed descriptions of the extracted time series features and the preprocessing transformations are provided in Appendix~\ref{appendix:features} and Appendix~\ref{appendix:preprocessing}. Hardware specifications and complete model hyperparameters, including training configurations and lookback window settings, are documented in Appendix~\ref{appendix:hardware} and Appendix~\ref{appendix:hyperparameters}. Mathematical definitions for all evaluation metrics are listed in Appendix~\ref{appendix:metrics}. Finally, we provide anonymized links to our source code and the complete generated metadataset in the abstract. This allows researchers to reproduce all experiments, figures, and tables presented in this work.

\bibliography{iclr2027_conference}
\bibliographystyle{iclr2027_conference}

\appendix
\section{Limitations and Future Works}

This study was designed to isolate the effect of preprocessing on time series forecasting benchmarks. Consequently, several design choices should be interpreted as deliberate controls rather than exhaustive coverage of all possible forecasting settings.

First, the preprocessing search space was restricted to a small set of established, interpretable, and reversible transformations. This design makes the benchmark controlled and reproducible, while allowing performance differences to be more directly attributed to the preprocessing pipeline. However, it excludes many other preprocessing strategies that may be relevant in practice, such as moving-average smoothing, learned normalization, and decomposition-based methods. Future work could expand the benchmark to include a richer set of preprocessing techniques, while explicitly distinguishing between reversible transformations, lossy smoothing operations, and representation-changing transformations.

Second, all experiments were conducted on univariate time series. This choice was made to avoid confounding preprocessing effects with cross-variable dependencies, covariate selection, or the ability of a model to extract information from exogenous variables. This is also why decomposition-based methods were not considered. However, many real-world forecasting problems are multivariate, and preprocessing may interact differently with models when multiple correlated variables are available. Future work should therefore extend the benchmark to multivariate forecasting tasks and examine whether the conclusions about structural preprocessing bias remain stable when models can use covariates or multiple target-related signals. A similar extension would be valuable for long-term forecasting settings, where preprocessing choices may affect error accumulation and temporal representation in different ways. Additionally, while our evaluation utilizes the M4 dataset, we emphasize that this corpus provides a rigorous, large-scale foundation for analyzing preprocessing sensitivity. Generating $\sim 5.1$ million experimental evaluations, prior PCA and t-SNE analyses~\citep{m4pca, gratis} have demonstrated that M4 captures wide variations in autocorrelation, trend strength, spectral entropy, and seasonality, representing a broad and realistic feature space.

Third, model hyperparameters were largely fixed to make preprocessing the main experimental variable. This improves interpretability because performance changes can be more directly associated with preprocessing choices rather than model tuning. However, it also means that the reported rankings should not be interpreted as a fully optimized leaderboard. Some models may benefit more than others from hyperparameter search, and the best preprocessing pipeline for a model may change when its architecture, learning rate, context length, or regularization parameters are tuned. A future preprocessing-aware leaderboard should therefore include model-specific hyperparameter optimization, while still controlling for fairness and computational budget.

Finally, the release of the PATM lays the groundwork for future advancements in meta-learning. By providing millions of evaluations across diverse models and pipelines, PATM enables researchers to train meta-models that recommend optimal preprocessing strategies based on time series meta-features. We view this as a crucial step toward fully automated, data-driven forecasting pipelines that dynamically adapt to the statistical structure of any given series.

%\section{Time Series Preliminaries}

\section{Benchmark Literature Review}
\label{appendix:benchmarks}

Robust and standardized benchmarks are essential for evaluating the performance of time series forecasting models, enabling fair comparison and reproducible research. Historically, large-scale competitions such as M4 and M5~\citep{m4competition} have served as crucial benchmarks, providing extensive collections of diverse time series. Building upon these foundations, the Monash Time Series Forecasting Archive~\citep{benchmark:monash} expanded the scope by aggregating datasets across varied domains and frequencies. Recent efforts have focused on providing unified toolkits to handle the increasing complexity of forecasting models. For instance, TFB~\citep{benchmark:tfb} introduces a scalable suite covering multiple domains, emphasizing reproducibility and aspect-based analysis. BasicTS+~\citep{benchmark:basicts} focuses on benchmarking multivariate forecasting by exploring spatial and temporal heterogeneity. Furthermore, with the rapid development of foundation models, benchmarks such as GIFT-Eval~\citep{benchmark:gifteval} and TSFM-Bench~\citep{benchmark:tsfm} have been designed specifically to evaluate zero-shot adaptation and prompt-based forecasting paradigms.

While the field of benchmarking for time series forecasting is continuously evolving to fill important gaps, a significant limitation persists across most frameworks. This limitation is the lack of a comprehensive and standardized preprocessing pipeline. This omission makes direct, fair comparisons between models challenging, as performance gains could be attributed to undisclosed preprocessing steps rather than architectural innovation. The difference in preprocessing pipelines between benchmarks can cause inconsistent results and add a bias toward models that are robust to different data structures.

Recent works have started addressing related limitations by exploring model design spaces and evaluating the effectiveness of specific components. For example, \citet{aretransformers} question the necessity of complex transformer architectures by demonstrating the strength of simple linear models under basic transformations. Similarly, TimeRecipe~\citep{timerecipe} explores the internal architectural design space of deep learning models by benchmarking time series forecasting methods at a finer-grained module level to uncover optimal configurations. Although these studies provide valuable insights into model and module selection, they do not yet directly analyze the impact of nonstationary preprocessing on different categories of models. They also do not explore how this external preprocessing phase impacts each category differently. This highlights the necessity of a framework that systematically evaluates the interaction between diverse forecasting algorithms and nonstationarity-handling pipelines.

\section{Time Series Features}
\label{appendix:features}

 The features stored for PATM provide a high-dimensional representation of the underlying dynamics of the data. These features extend the metadata originally distributed with the Monash Time Series Forecasting Archive~\citep{benchmark:monash} by incorporating additional statistical descriptors, particularly with \texttt{seasonal\_strength}, which quantifies the relative magnitude of seasonal patterns; \texttt{adf\_pvalue}, which captures the statistical significance of the Augmented Dickey--Fuller stationarity test; and \texttt{skewness}, which measures asymmetry in the empirical distribution of observations. The complete set of features was extracted using the \texttt{tsfeatures} package for Python, which implements a wide range of statistical procedures designed to summarize trend, seasonality, autocorrelation structure, complexity, and heteroscedasticity. Detailed descriptions for each of the 40 extracted features are provided in Table~\ref{tab:features}. The subset of features used for visualizations in this study is further described below.

\begin{longtable}{clp{10cm}}
\caption{Description of Time Series Features extracted via tsfeatures.}
\label{tab:features} \\
\toprule
\textbf{\#} & \textbf{Feature Name} & \textbf{Description} \\ \midrule
1 & mean & Mean value of the time series. \\
2 & var & Variance of the time series. \\
3 & max\_kl\_shift & Maximum Kullback-Leibler divergence detected across windows. \\
4 & time\_kl\_shift & Relative time index where the maximum KL shift occurs. \\
5 & max\_level\_shift & Maximum change in mean level between windows. \\
6 & time\_level\_shift & Relative time index where the maximum level shift occurs. \\
7 & max\_var\_shift & Maximum variance change between windows. \\
8 & time\_var\_shift & Relative time index where the maximum variance shift occurs. \\
9 & x\_acf1 & First autocorrelation coefficient of the original series. \\
10 & x\_acf10 & Sum of squared first 10 autocorrelations of original series. \\
11 & diff1\_acf1 & First autocorrelation of the first-differenced series. \\
12 & diff1\_acf10 & ACF sum of squares for the first-differenced series. \\
13 & diff2\_acf1 & First autocorrelation of the second-differenced series. \\
14 & diff2\_acf10 & ACF sum of squares for the second-differenced series. \\
15 & ARCH.LM & Lagrange Multiplier test statistic for ARCH effects. \\
16 & crossing\_points & Number of times the series crosses the median line. \\
17 & entropy & Spectral entropy measuring the complexity of the series. \\
18 & flat\_spots & Number of sections where the series is relatively constant. \\
19 & alpha & Smoothing parameter for the level in ETS models. \\
20 & beta & Smoothing parameter for the trend in ETS models. \\
21 & hurst & Hurst exponent measuring long-term memory or autocorrelation. \\
22 & lumpiness & Variance of the variances of tiled windows. \\
23 & nonlinearity & Teräsvirta's neural network test for non-linearity. \\
24 & x\_pacf5 & Sum of squares of the first 5 partial autocorrelations. \\
25 & diff1x\_pacf5 & PACF sum of squares for the first-differenced series. \\
26 & diff2x\_pacf5 & PACF sum of squares for the second-differenced series. \\
27 & stability & Variance of the means of tiled windows across the series. \\
28 & unitroot\_kpss & KPSS test statistic for stationarity of the series. \\
29 & unitroot\_pp & Phillips-Perron test statistic for presence of a unit root. \\
30 & nperiods & Number of seasonal periods identified in the data. \\
31 & seasonal\_period & Length of the primary seasonal cycle (e.g., 12 for monthly). \\
32 & trend & Strength of the trend component in the decomposition. \\
33 & spike & Variance of the error component in a leave-one-out analysis. \\
34 & linearity & Linearity of the trend component from the decomposition. \\
35 & curvature & Curvature (quadratic term) of the trend component. \\
36 & e\_acf1 & First partial autocorrelation of the remainder series. \\
37 & e\_acf10 & Sum of squared first 10 autocorrelations of the remainders. \\
38 & seasonal\_strength & Strength of the seasonal component in the decomposition. \\
39 & adf\_pvalue & P-value for the Augmented Dickey-Fuller test. \\
40 & skewness & Measure of the asymmetry of the data distribution. \\ \bottomrule
\end{longtable}

\textbf{Trend and Season.} \textit{Trend} describes the long-term direction of movement in a time series, while \textit{seasonality} refers to systematic patterns that repeat over a fixed period. Using Seasonal and Trend decomposition based on Loess (STL)~\citep{stl}, a series can be represented as
\[
x_t = T_t + S_t + R_t,
\]
where \(T_t\) is the trend component, \(S_t\) is the seasonal component, and \(R_t\) is the remainder. Based on this decomposition, the strength of trend and seasonality can be measured as
\[
F_T = \max\left(0,\;1-\frac{\mathrm{Var}(R_t)}{\mathrm{Var}(x_t-S_t)}\right)
\]
and
\[
F_S = \max\left(0,\;1-\frac{\mathrm{Var}(R_t)}{\mathrm{Var}(x_t-T_t)}\right),
\]
respectively. Values closer to 1 indicate a stronger trend or seasonality, while values closer to 0 indicate a weaker structure. In this study, high strengths for seasonality and trend is defined as values exceeding 0.75, while low strength corresponds to values below 0.25.

\textbf{Stationarity.} A series is considered \textit{strictly stationary} if, for any time shifts $h\in \mathbb{Z}$, and any arbitrary values $a_1, a_2, \dots, a_n$,

\begin{equation}\label{eq:strictstationary}
P(x_{1} \le a_1,\dots ,x_{n} \le a_n) \;=\; P(x_{1+h} \le a_1, \dots, x_{n+h} \le a_n).
\end{equation}

That is, all finite-dimensional joint distributions are invariant to time shifts.

A process is \textit{weakly stationary} if it has finite second moments and (i) its mean function $E(x_t)$ is constant and does not depend on any time index $t$, and (ii) the covariance of an arbitrary $x_t$ and time shifted $x_{t+h}$ depends only on $h$. Weak stationarity is generally sufficient for standard time series modeling~\citep{timeseriesbook}. \textit{Unless stated otherwise, “stationary” in this report refers to weak stationarity.} To assess this property, we employ the Augmented Dickey--Fuller (ADF) test~\citep{adf}, and the decision rule is given in Equation~\ref{eq:stationarity}.

\begin{equation}
\label{eq:stationarity}
\mathrm{Stationarity}
=
\begin{cases}
\mathrm{True}, & \mathrm{ADF}(X) \le 0.05, \\
\mathrm{False}, & \mathrm{ADF}(X) > 0.05.
\end{cases}
\end{equation}

\textbf{Entropy.} Spectral entropy measures the complexity or predictability of a time series. It computes the Shannon entropy of the normalized spectral density of the series. Values closer to 1 indicate a highly complex and unpredictable series (similar to white noise), whereas values closer to 0 indicate a more predictable series with strong structural patterns. We classify series with an entropy greater than 0.75 as high entropy and those below 0.25 as low entropy.

\textbf{Shifting.} Shifting captures abrupt structural breaks in the time series, which can manifest as sudden changes in the mean level, variance, or overall distribution. To quantify shifting, we aggregate features such as the maximum Kullback-Leibler divergence (\texttt{max\_kl\_shift}), maximum level shift (\texttt{max\_level\_shift}), and maximum variance shift (\texttt{max\_var\_shift}) detected across rolling windows. Series with significant breaks are categorized as high shifting, while those with stable properties are considered low shifting.

\section{Categories of Preprocessing}
\label{appendix:preprocessing}

Here, $w_t$ denotes the transformed series that is closer to a weak stationary series. The design of our experimental groups was constructed specifically to disentangle the core components of non-stationarity, such as variance, trend, and seasonality, and observe how their progressive removal interacts with different architectures. Operations follow a strict mathematical sequence to ensure validity, for example testing scaling in isolation versus combinations of variance stabilization and differencing.

\textbf{Variance-Stabilizing Transforms.} Variance-stabilizing transforms shrink large values more than small ones, helping residual variance look roughly constant over time. To stabilize variance that evolves with the mean level of the series, we employ the Box-Cox Transformation~\citep{boxcoxtransform}. This power transformation is a generalization of the logarithmic transform and is defined as:

\begin{equation}\label{eq:log}
w_t = 
\begin{cases} 
\frac{x_t^\lambda - 1}{\lambda} & \text{if } \lambda \neq 0 \\ 
\ln(x_t) & \text{if } \lambda = 0 
\end{cases}
\end{equation}

where $\lambda$ is the transformation parameter. When $\lambda = 0$, the transformation is equivalent to the natural logarithm, which constrains forecasts to be positive on the original scale. By varying $\lambda$, the Box-Cox transform can address a wide range of heteroscedastic behaviors, ensuring the residual variance remains approximately constant over time.

%\subsubsection{Difference-based Unit Root Removal}

\textbf{Difference-based Unit Root Removal.} Difference-based transforms target mean stability by removing changes in level and, with seasonal lags, strong seasonal patterns. A first-order differenced series (Equation~\ref{eq:firstorderdiff}) has the difference between consecutive values from the original series, removing linear deterministic trends and converting stochastic trends into a stationary series under standard conditions.

\begin{equation}\label{eq:firstorderdiff}
w_t \;=\; x_t - x_{t-1}.
\end{equation}

When seasonality of period $m$ induces a seasonal unit root, seasonal differencing can convert it into a stationary series by taking the $m$-seasonal difference.

\begin{equation}\label{eq:sdiff}
w_t \;=\; x_t - x_{t-m}.
\end{equation}

Forecasts produced on the differenced scale must be re-integrated to the original scale by summation of previously differenced values.

%\subsubsection{Scaling}

\textbf{Scaling.} Scaling adjusts the magnitude of the data to ensure that features contribute equally to the model objective function. While this step is standard, it plays an important role in practice, as many optimization procedures are sensitive to the scale of the inputs, and unscaled data can lead to slower convergence or poorly conditioned objective functions. The transformation is defined as follows:

\begin{equation}\label{eq:scaling}
w_t = \frac{x_t - \mu}{\sigma},
\end{equation}

where $\mu$ represents the mean and $\sigma$ represents the standard deviation of the series. For forecasting applications, these parameters are typically estimated from the training set to prevent data leakage from the test set. Similar to other preprocessing steps, the scaling process is invertible, allowing the predicted values to be rescaled to the original units for performance evaluation and interpretation.

This research employs Z-Score Normalization~\citep{scikit}, which transforms the series to have a mean of zero and a unit variance.

% \begin{table}[h!]
% \centering
% \small
% \setlength{\tabcolsep}{6pt}
% \begin{tabular}{lp{0.7\textwidth}}
% \toprule
% \textbf{Preprocessing Step} & \textbf{Models} \\
% \midrule
% Variance-Stabilizing 
% & ETS~\cite{ets}, TBATS~\cite{tbats} \\

% Difference-Based 
% & ETS~\cite{ets}, ARIMA~\cite{timeseriesbook}, TBATS~\cite{tbats} \\

% Detrending 
% & ETS~\cite{ets}, ARIMA~\cite{timeseriesbook}, THETA~\cite{theta}, TBATS~\cite{tbats} \\

% Decomposition 
% & ETS~\cite{ets}, THETA~\cite{theta}, TBATS~\cite{tbats}, FEDformer~\cite{fedformer} \\

% Scaling / Normalization 
% & TimesNet~\cite{timesnet}, PatchTST~\cite{patchtst}, FEDformer~\cite{fedformer}, Informer~\cite{informer}, Crossformer~\cite{crossformer} \\

% \bottomrule
% \end{tabular}
% \caption{Preprocessing steps embedded in model architectures. Models are grouped by the preprocessing operations they internally implement. \textcolor{red}{I want to remake this Table, but more directly mentioning IP-Models}}
% \label{tab:model_preprocessing}
% \end{table}

\section{Hardware Specifications}
\label{appendix:hardware}

All experiments were conducted on a server equipped with an \textbf{Intel Xeon Silver 4310 CPU (2.10GHz)}, \textbf{512GB DDR4 RAM}, and an \textbf{NVIDIA RTX A6000 (48GB VRAM)}. Classical statistical models were implemented using \textit{statsforecast}, while machine learning and deep learning models utilized \textit{scikit-learn}, \textit{XGBoost}, and \textit{PyTorch}. 

\section{Model Hyperparameters and Training Configurations}
\label{appendix:hyperparameters}

While deep learning models could undoubtedly achieve better absolute accuracy with individualized tuning, our explicit objective in this benchmark is to evaluate the isolated impact of data preprocessing across different model families. By adopting fixed model hyperparameters and lag settings across all experiments based on established conventions, we strictly isolate preprocessing as the single independent variable. This ensures that any observed performance shifts are attributable entirely to pipeline conditioning rather than confounding variations in capacity. Furthermore, while these standardized lookback windows (input lengths) were used to maintain experimental control, they were not chosen arbitrarily. They were adopted directly from the specifications documented by the Monash Time Series Forecasting Archive~\citep{benchmark:monash}. Overall, these specifications systematically align the input window length with the temporal frequency and dominant seasonal period of the dataset to provide a consistent evaluation baseline. As detailed earlier in Table~\ref{tab:m4_characteristics}, this translates to lookback windows that capture the fundamental seasonal cycle, such as 24 steps for hourly series or 7 steps for daily series. It is important to note that input length does interact with preprocessing. For example, seasonal differencing requires an input length strictly longer than the seasonal period $m$. Additionally, deep learning models with fixed context windows might receive slightly different temporal dynamics if aggressive differencing truncates the earliest observations. A joint exploration of hyperparameter optimization, input lengths, and preprocessing sensitivity remains an important direction for future research.

\subsection{Statistical Models}

For ETS, ARIMA, and Theta, we utilize automated selection procedures to determine the optimal error/trend/seasonal components, lag orders, and smoothing parameters, respectively. These are implemented via the \textit{statsforecast} library using seasonal strength detection. Automated selection for statistical algorithms is a well-established practice in benchmark studies~\citep{benchmark:monash, benchmark:tfb, benchmark:gifteval, benchmark:tsfm}.

\subsection{Machine Learning Baselines}

The machine learning baselines, Linear Regression and XGBoost, are configured following standard forecasting benchmarks.

\begin{table}[ht]
\caption{Hyperparameters for Machine Learning Baselines}
\centering
\begin{tabular}{@{}lll@{}}
\toprule
\textbf{Hyperparameter} & \textbf{Linear Regression} & \textbf{XGBoost} \\ \midrule
Library & scikit-learn & XGBoost \\
Max Depth & N/A & 6 \\
Estimators & N/A & 1000 \\
Subsample & N/A & 0.8 \\
Colsample bytree & N/A & 0.8 \\
L2 Regularization ($\lambda$) & 1.0 & 1.0 \\ \bottomrule
\end{tabular}
\label{tab:ml_hyperparams}
\end{table}

\subsection{Deep Learning and Transformer Architectures}

To maintain consistency, neural models were configured such that their capacity is roughly aligned, with the exception of Transformer baselines which follow the recommended configurations from their original publications.

\begin{table}[ht]
\caption{Hyperparameters for Deep Learning Models}
\centering
\begin{tabular}{@{}llll@{}}
\toprule
\textbf{Hyperparameter} & \textbf{MLP} & \textbf{N-BEATS} & \textbf{TimesNet} \\ \midrule
Architecture & 2 Hidden Layers & 6 Stacks / 4 Layers & 2 Encoder Layers \\
Hidden Dim ($d_{model}$) & 512 & 256 (Layer Size) & 16 \\
FFN Dim ($d_{ff}$) & N/A & N/A & 32 \\
Activation & ReLU & ReLU & GeLU \\ \bottomrule
\end{tabular}
\label{tab:dl_hyperparams}
\end{table}

\subsection{Transformer-based Architectures}

We include PatchTST and Informer as representatives of state-of-the-art attention-based modeling. Their hyperparameters follow the standard configurations suggested in their respective original implementations.

\begin{table}[ht]
\caption{Hyperparameters for Transformer-based Models}
\centering
\begin{tabular}{@{}lll@{}}
\toprule
\textbf{Hyperparameter} & \textbf{PatchTST} & \textbf{Informer} \\ \midrule
Layers & 3 Encoder & 2 Encoder / 1 Decoder \\
Attention Heads & 16 & 4 \\
Model Dim ($d_{model}$) & 128 & 256 \\
FFN Dim ($d_{ff}$) & 256 & 512 \\
Activation & GeLU & GeLU \\
Normalization & RevIN & Standard \\ \bottomrule
\end{tabular}
\label{tab:transformer_hyperparams}
\end{table}

\section{Local vs. Global Models}
\label{appendix:local_global}

In time series forecasting, machine learning and deep learning architectures can be trained using two distinct paradigms, local and global~\citep{benidis2022deep}. A local model is trained exclusively on a single time series and predicts only for that specific series. Conversely, a global model is trained jointly across an entire dataset or a large collection of time series. This global approach allows the model to learn shared patterns and parameters that apply across all instances.

We decided to run our machine learning and deep learning models in a global setup for our primary evaluation. This approach aligns directly with standard practices in contemporary forecasting literature~\citep{patchtst}, where large capacity architectures are typically trained globally to leverage cross-series information and maximize data availability.

\begin{figure}[h] \centering \includegraphics[width=0.75\textwidth]{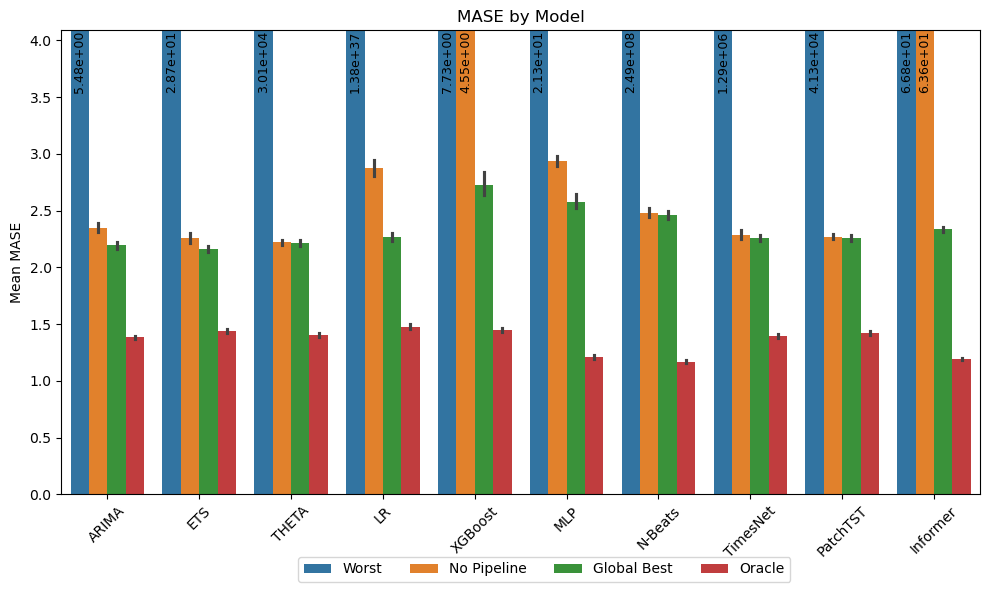} \caption{Barplot MASE of best and oracle pipelines of local models. Lower loss values mean better results.} \label{fig:oracle_mase_local} \end{figure}

However, to strictly isolate the per-series impact of preprocessing pipelines without the potential interference of cross-series context sharing, we also ran all our experiments locally. The core results and interpretations remained fundamentally the same across both paradigms. Under both settings, models without internalized preprocessing continued to exhibit strong sensitivity to external pipelines, while models with internalized mechanisms showed more stable relative performance. The results currently appearing as Figure~\ref{fig:oracle_mase_local} and Table~\ref{tab:local_results} correspond to the experiments evaluated as local models.

\begin{table}[ht]
\caption{Forecasting performance (MASE) across preprocessing pipelines and models when trained on each series locally. Best per model in bold. Columns in grey are IP-Models. Pipelines are composed of: \textbf{X} (Baseline/None), \textbf{V} (Boxcox transformation), \textbf{U} (First-order differencing), \textbf{S} (Seasonal differencing), and \textbf{Q} (Standard scaling).}

\centering
\small
\resizebox{0.8\columnwidth}{!}{
\begin{tabularx}{\textwidth}{l G G G N N N N G G N}
\toprule
Pipeline & ARIMA & ETS & THETA & LR & XGBst & MLP & NBTS & TNet & PTST & Informer \\
\midrule
X    & 2.35 & 2.25 & 2.22 & 2.89 & 4.55 & 2.94 & \textbf{2.48} & 2.29 & 2.27 & 63.59 \\
V    & 3.19 & 2.38 & 2.36 & 3.77 & 4.54 & 6.42 & 2.49 & \textbf{2.26} & \textbf{2.26} & 18.23 \\
U    & 2.23 & 2.40 & 2.60 & \textbf{2.28} & \textbf{2.73} & 3.01 & 3.02 & 3.27 & 3.08 & \textbf{2.35} \\
S    & 2.38 & 3.19 & 3.19 & 2.71 & 3.26 & 3.47 & 3.59 & 3.35 & 3.10 & 3.04 \\
Q    & 2.30 & \textbf{2.18} & \textbf{2.22} & 2.89 & 4.55 & 3.97 & 3.76 & 2.27 & 2.27 & 4.95 \\
VU   & 2.23 & 2.45 & 2.84 & 2.45 & 2.87 & \textbf{2.58} & 3.19 & 3.83 & 3.39 & 4.74 \\
VS   & 2.81 & 3.85 & 3.82 & $2.35e^{4}$ & 3.44 & 6.01 & 3.28 & 3.74 & 3.28 & 3.88 \\
VQ   & 2.27 & 2.38 & 2.38 & 3.77 & 4.54 & 4.59 & 6.04 & 2.27 & \textbf{2.26} & 5.25 \\
US   & 2.44 & 3.81 & 4.47 & 3.13 & 3.69 & 4.03 & 4.17 & 6.95 & 6.55 & 3.24 \\
UQ   & \textbf{2.20} & 2.34 & 2.60 & \textbf{2.28} & \textbf{2.73} & 3.00 & 3.08 & 3.27 & 3.08 & 3.03 \\
SQ   & 2.39 & 3.13 & 3.19 & 2.71 & 3.26 & 3.61 & 3.51 & 3.36 & 3.10 & 3.33 \\
VUS  & 2.57 & $2.70e^{1}$ & $2.99e^{4}$ & $1.38e^{37}$ & 4.42 & 6.39 & $2.49e^{8}$ & $1.23e^{6}$ & $3.71e^{4}$ & 7.10 \\
VUQ  & 2.33 & 2.45 & 2.85 & 2.45 & 2.87 & 3.21 & 3.44 & 3.62 & 3.35 & 3.33 \\
VSQ  & 2.91 & 3.59 & 3.83 & $2.35e^{4}$ & 3.44 & 9.32 & 4.23 & 3.59 & 3.29 & 3.71 \\
USQ  & 2.72 & 3.65 & 4.47 & 3.13 & 3.69 & 4.11 & 4.26 & 6.90 & 6.54 & 4.34 \\
VUSQ & 3.22 & $2.70e^{1}$ & $3.01e^{4}$ & $1.38e^{37}$ & 4.42 & 4.86 & $1.75e^{2}$ & $8.92e^{4}$ & $3.29e^{4}$ & 5.84 \\
\bottomrule
\end{tabularx}
}
\label{tab:local_results}
\end{table}

\section{Evaluation Metrics}
\label{appendix:metrics}

These metrics capture different aspects of predictive accuracy, such as sensitivity to large deviations, scale dependence, and robustness to varying magnitudes across series.

Let $y_i$ denote the true value, $\hat{y}_i$ the predicted value, and $N$ the total number of evaluated predictions. 

\subsection{Mean Absolute Scaled Error (MASE)}

Mean Absolute Scaled Error (MASE) measures the absolute forecasting error relative to the error of a naive baseline. For a seasonal time series with seasonality period $m$, the scaling term is computed from the training sequence $\{y_t\}_{t=1}^{T}$ as
\begin{equation}
Q
=
\frac{1}{T-m}
\sum_{t=m+1}^{T}
\left| y_t - y_{t-m} \right| .
\end{equation}
MASE is then defined as
\begin{equation}
\mathrm{MASE}
=
\frac{1}{N}
\sum_{i=1}^{N}
\frac{
\left| y_i - \hat{y}_i \right|
}{
Q + \epsilon
},
\end{equation}
where $\epsilon$ is a small positive constant for numerical stability. When $m=1$, the denominator corresponds to the average one-step naive forecasting error.

\subsection{Mean Squared Error (MSE)}

Mean Squared Error (MSE) measures the average squared deviation between the predicted values and the ground-truth observations. MSE is defined as
\begin{equation}
\mathrm{MSE}
=
\frac{1}{N}
\sum_{i=1}^{N}
\left(y_i - \hat{y}_i\right)^2 .
\end{equation}
Because the errors are squared, MSE penalizes large forecasting errors more heavily than small ones. Lower MSE indicates better predictive accuracy.

\begin{table}[ht]
\caption{Forecasting performance (MSE) across preprocessing pipelines and models. Best per model in bold. Columns in grey are IP-Models. Pipelines are composed of: \textbf{X} (Baseline/None), \textbf{V} (Boxcox transformation), \textbf{U} (First-order differencing), \textbf{S} (Seasonal differencing), and \textbf{Q} (Standard scaling).}
\centering
\small

\begin{tabularx}{\textwidth}{l G G G N N N N G G N G}

\toprule
Pipeline & ARIMA & ETS & THETA & LR & XGBst & MLP & NBTS & TNet & PTST & Informer & Chronos \\
\midrule
X    & $1.94e^{6}$ & $9.96e^{6}$ & $\bm{1.63e^{6}}$ & $7.07e^{6}$ & $2.21e^{6}$ & $3.23e^{7}$ & $3.10e^{6}$ & $1.95e^{6}$ & $1.93e^{6}$ & $5.01e^{7}$ & $1.60e^{6}$ \\
V    & $2.35e^{6}$ & $2.14e^{6}$ & $1.91e^{6}$ & $9.16e^{8}$ & $2.81e^{6}$ & $1.19e^{7}$ & $3.14e^{9}$ & $1.78e^{6}$ & $1.92e^{6}$ & $6.26e^{7}$ & $1.73e^{6}$ \\
U    & $1.78e^{6}$ & $5.17e^{6}$ & $4.76e^{6}$ & $\bm{2.40e^{6}}$ & $1.79e^{6}$ & $2.76e^{6}$ & $8.88e^{7}$ & $4.63e^{6}$ & $6.29e^{6}$ & $2.62e^{6}$ & $1.96e^{6}$ \\
S    & $2.00e^{6}$ & $3.82e^{6}$ & $3.99e^{6}$ & $1.90e^{7}$ & $1.83e^{6}$ & $2.18e^{6}$ & $5.23e^{6}$ & $3.65e^{6}$ & $4.85e^{6}$ & $2.10e^{6}$ & $2.24e^{6}$ \\
Q    & $\bm{1.71e^{6}}$ & $3.53e^{6}$ & $1.64e^{6}$ & $7.07e^{6}$ & $2.85e^{6}$ & $3.04e^{6}$ & $4.07e^{6}$ & $1.77e^{6}$ & $1.95e^{6}$ & $3.81e^{6}$ & $\bm{1.60e^{6}}$ \\
VU   & $2.01e^{6}$ & $2.93e^{6}$ & $3.01e^{7}$ & $5.45e^{6}$ & $\bm{1.72e^{6}}$ & $2.18e^{6}$ & $1.18e^{58}$ & $7.98e^{8}$ & $9.21e^{8}$ & $2.53e^{6}$ & $2.14e^{6}$ \\
VS   & $1.23e^{8}$ & $8.83e^{11}$ & $1.03e^{9}$ & $5.34e^{20}$ & $1.86e^{6}$ & $1.98e^{6}$ & $3.41e^{6}$ & $2.87e^{8}$ & $2.77e^{8}$ & $2.16e^{6}$ & $2.76e^{7}$ \\
VQ   & $1.93e^{6}$ & $\bm{2.14e^{6}}$ & $1.88e^{6}$ & $9.16e^{8}$ & $2.93e^{6}$ & $3.68e^{6}$ & $2.39e^{6}$ & $\bm{1.72e^{6}}$ & $\bm{1.85e^{6}}$ & $3.95e^{6}$ & $1.71e^{6}$ \\
US   & $2.23e^{6}$ & $7.01e^{6}$ & $1.12e^{7}$ & $1.96e^{7}$ & $3.63e^{6}$ & $5.23e^{6}$ & $1.08e^{8}$ & $2.48e^{7}$ & $2.47e^{7}$ & $5.02e^{6}$ & $6.46e^{6}$ \\
UQ   & $1.86e^{6}$ & $3.92e^{6}$ & $4.76e^{6}$ & $2.40e^{6}$ & $1.79e^{6}$ & $1.99e^{6}$ & $8.09e^{7}$ & $5.33e^{6}$ & $5.73e^{6}$ & $2.34e^{6}$ & $1.95e^{6}$ \\
SQ   & $2.03e^{6}$ & $3.85e^{6}$ & $3.98e^{6}$ & $1.90e^{7}$ & $1.85e^{6}$ & $\bm{1.96e^{6}}$ & $\bm{2.13e^{6}}$ & $3.65e^{6}$ & $3.63e^{6}$ & $\bm{2.01e^{6}}$ & $2.24e^{6}$ \\
VUS  & $3.04e^{7}$ & $5.66e^{11}$ & $6.19e^{18}$ & $1.93e^{86}$ & $7.25e^{7}$ & $1.80e^{7}$ & $1.64e^{81}$ & $9.06e^{20}$ & $2.10e^{22}$ & $2.06e^{8}$ & $7.87e^{13}$ \\
VUQ  & $2.39e^{6}$ & $2.94e^{6}$ & $2.87e^{7}$ & $5.45e^{6}$ & $2.43e^{6}$ & $2.87e^{6}$ & $1.03e^{58}$ & $8.89e^{8}$ & $1.76e^{9}$ & $3.38e^{6}$ & $2.14e^{6}$ \\
VSQ  & $1.17e^{8}$ & $7.29e^{8}$ & $1.01e^{9}$ & $5.34e^{20}$ & $2.78e^{6}$ & $2.58e^{6}$ & $1.57e^{9}$ & $2.91e^{8}$ & $2.28e^{8}$ & $2.45e^{6}$ & $2.76e^{7}$ \\
USQ  & $2.96e^{6}$ & $6.71e^{6}$ & $1.12e^{7}$ & $1.96e^{7}$ & $3.57e^{6}$ & $3.81e^{6}$ & $4.71e^{6}$ & $2.60e^{7}$ & $2.59e^{7}$ & $4.82e^{6}$ & $6.46e^{6}$ \\
VUSQ & $1.75e^{9}$ & $5.66e^{11}$ & $1.36e^{19}$ & $1.93e^{86}$ & $1.23e^{9}$ & $4.92e^{8}$ & $1.78e^{81}$ & $4.12e^{26}$ & $2.69e^{22}$ & $4.97e^{9}$ & $7.92e^{13}$ \\
\bottomrule
\end{tabularx}
\label{tab:main_results_mse}
\end{table}

\subsection{Root Mean Squared Error (RMSE)}

Root Mean Squared Error (RMSE) is the square root of MSE. It is defined as
\begin{equation}
\mathrm{RMSE}
=
\sqrt{
\frac{1}{N}
\sum_{i=1}^{N}
\left(y_i - \hat{y}_i\right)^2
}.
\end{equation}
Unlike MSE, RMSE is expressed in the same unit as the target variable, which makes it easier to interpret. Similar to MSE, RMSE assigns relatively large penalties to large errors.

\begin{table}[ht]
\caption{Forecasting performance (RMSE) across preprocessing pipelines and models. Best per model in bold. Columns in grey are IP-Models. Pipelines are composed of: \textbf{X} (Baseline/None), \textbf{V} (Boxcox transformation), \textbf{U} (First-order differencing), \textbf{S} (Seasonal differencing), and \textbf{Q} (Standard scaling).}
\centering
\small

\begin{tabularx}{\textwidth}{l G G G N N N N G G N G}

\toprule
Pipeline & ARIMA & ETS & THETA & LR & XGBst & MLP & NBTS & TNet & PTST & Informer & Chronos \\
\midrule
X    & 618.42 & 648.65 & \textbf{591.69} & 721.70 & 661.90 & 1423.34 & 724.41 & 670.77 & 689.61 & 3891.09 & 593.42 \\
V    & 674.92 & 610.30 & 615.76 & 949.83 & 685.92 & 786.98 & 1243.25 & \textbf{642.18} & \textbf{652.60} & 4349.75 & 624.72 \\
U    & 602.66 & 661.07 & 736.04 & \textbf{621.58} & 615.27 & 680.22 & 2082.37 & 882.41 & 942.62 & 680.39 & 593.82 \\
S    & 624.96 & 804.39 & 821.39 & 733.01 & 629.87 & 718.31 & 887.38 & 806.01 & 919.89 & 706.19 & 657.58 \\
Q    & 610.05 & \textbf{608.39} & 594.12 & 721.61 & 854.76 & 923.15 & 1025.95 & 652.68 & 672.63 & 1026.64 & \textbf{593.26} \\
VU   & 607.93 & 667.51 & 815.84 & 668.15 & \textbf{604.06} & 685.63 & $6.40e^{26}$ & 1446.61 & 1473.95 & 711.34 & 614.74 \\
VS   & 707.48 & 6815.38 & 1458.23 & $1.36e^{8}$ & 621.78 & 656.73 & 673.55 & 1092.80 & 1106.79 & 695.86 & 769.08 \\
VQ   & 613.73 & 610.16 & 621.09 & 949.86 & 863.52 & 1004.09 & 707.33 & 648.06 & 664.09 & 1010.87 & 606.99 \\
US   & 637.93 & 991.67 & 1231.19 & 842.53 & 759.16 & 884.45 & 3071.64 & 1797.69 & 1781.70 & 931.80 & 890.15 \\
UQ   & \textbf{599.66} & 647.81 & 736.40 & 621.63 & 615.74 & \textbf{631.41} & 1065.32 & 922.70 & 898.47 & 660.90 & 593.80 \\
SQ   & 625.38 & 799.79 & 820.88 & 733.13 & 622.24 & 645.38 & \textbf{660.50} & 799.74 & 795.33 & \textbf{649.58} & 657.59 \\
VUS  & 712.00 & 7559.64 & $1.51e^{7}$ & $8.16e^{40}$ & 941.81 & 903.38 & $2.38e^{38}$ & $2.57e^{8}$ & $1.07e^{9}$ & 1137.14 & $5.44e^{4}$ \\
VUQ  & 634.77 & 667.80 & 816.67 & 668.17 & 653.22 & 701.44 & $5.98e^{26}$ & 1405.64 & 1744.87 & 699.24 & 614.72 \\
VSQ  & 738.15 & 1293.15 & 1452.54 & $1.36e^{8}$ & 664.56 & 682.76 & 1682.53 & 1113.68 & 1020.35 & 689.45 & 769.08 \\
USQ  & 709.20 & 977.98 & 1229.52 & 842.55 & 787.41 & 807.54 & 882.53 & 1793.12 & 1783.23 & 864.65 & 890.15 \\
VUSQ & 1222.50 & 7565.95 & $3.07e^{7}$ & $8.16e^{40}$ & 1258.40 & 1073.74 & $2.48e^{38}$ & $1.21e^{11}$ & $1.26e^{9}$ & 1528.40 & $5.46e^{4}$ \\
\bottomrule
\end{tabularx}
\label{tab:main_results_rmse}
\end{table}

\subsection{Mean Absolute Error (MAE)}

Mean Absolute Error (MAE) measures the average absolute deviation between predictions and observations. It is defined as
\begin{equation}
\mathrm{MAE}
=
\frac{1}{N}
\sum_{i=1}^{N}
\left| y_i - \hat{y}_i \right| .
\end{equation}
Compared with MSE and RMSE, MAE is less sensitive to extreme errors because it does not square the residuals. Lower MAE indicates better forecasting performance.

\begin{table}[ht]
\caption{Forecasting performance (MAE) across preprocessing pipelines and models. Best per model in bold. Columns in grey are IP-Models. Pipelines are composed of: \textbf{X} (Baseline/None), \textbf{V} (Boxcox transformation), \textbf{U} (First-order differencing), \textbf{S} (Seasonal differencing), and \textbf{Q} (Standard scaling).}
\centering
\small

\begin{tabularx}{\textwidth}{l G G G N N N N G G N G}

\toprule
Pipeline & ARIMA & ETS & THETA & LR & XGBst & MLP & NBTS & TNet & PTST & Informer & Chronos \\
\midrule
X    & 526.24 & 553.24 & \textbf{504.76} & 614.33 & 569.67 & 1324.94 & 620.81 & 574.60 & 591.64 & 3833.70 & 502.59 \\
V    & 581.59 & 518.54 & 524.29 & 731.96 & 589.24 & 694.80 & 852.16 & \textbf{550.80} & \textbf{559.18} & 4296.38 & 531.81 \\
U    & 512.02 & 564.21 & 633.66 & \textbf{526.75} & 525.10 & 580.70 & 1842.06 & 766.06 & 823.37 & 581.06 & 505.46 \\
S    & 531.41 & 686.34 & 702.94 & 618.88 & 537.41 & 618.67 & 767.10 & 699.94 & 812.81 & 603.85 & 557.67 \\
Q    & 518.92 & \textbf{517.77} & 507.36 & 614.21 & 745.04 & 826.66 & 927.45 & 558.95 & 578.99 & 932.19 & \textbf{502.46} \\
VU   & 514.37 & 570.43 & 689.56 & 562.35 & \textbf{515.75} & 589.72 & $1.18e^{26}$ & 1134.40 & 1154.18 & 612.51 & 523.73 \\
VS   & 586.07 & 2994.63 & 1139.41 & $4.90e^{7}$ & 529.56 & 561.19 & 572.47 & 911.62 & 921.35 & 596.10 & 644.08 \\
VQ   & 519.82 & 518.54 & 530.25 & 731.99 & 751.82 & 872.34 & 605.18 & 555.68 & 569.45 & 903.31 & 516.90 \\
US   & 543.87 & 852.04 & 1055.15 & 717.05 & 652.92 & 764.74 & 2591.42 & 1540.06 & 1522.94 & 807.85 & 766.35 \\
UQ   & \textbf{511.50} & 553.24 & 633.94 & 526.80 & 525.82 & \textbf{541.55} & 919.15 & 802.22 & 780.65 & 564.33 & 505.44 \\
SQ   & 534.08 & 682.90 & 702.50 & 619.00 & 531.62 & 553.43 & \textbf{563.36} & 693.42 & 689.04 & \textbf{554.34} & 557.67 \\
VUS  & 602.11 & 3897.35 & $5.46e^{6}$ & $2.88e^{40}$ & 786.00 & 768.82 & $3.56e^{37}$ & $9.41e^{7}$ & $3.89e^{8}$ & 920.59 & $2.24e^{4}$ \\
VUQ  & 541.44 & 570.68 & 690.55 & 562.37 & 559.43 & 605.23 & $1.10e^{26}$ & 1103.70 & 1334.46 & 598.49 & 523.71 \\
VSQ  & 617.86 & 1030.74 & 1136.64 & $4.90e^{7}$ & 567.77 & 584.37 & 1282.96 & 927.82 & 856.77 & 588.80 & 644.10 \\
USQ  & 608.37 & 841.38 & 1053.78 & 717.07 & 678.21 & 698.94 & 759.87 & 1535.05 & 1526.92 & 748.18 & 766.34 \\
VUSQ & 947.37 & 3900.43 & $1.11e^{7}$ & $2.89e^{40}$ & 981.56 & 867.44 & $3.71e^{37}$ & $4.36e^{10}$ & $4.59e^{8}$ & 1121.99 & $2.25e^{4}$ \\
\bottomrule
\end{tabularx}
\label{tab:main_results_mae}
\end{table}

\subsection{Symmetric Mean Absolute Percentage Error (sMAPE)}

Symmetric Mean Absolute Percentage Error (sMAPE) evaluates the prediction error relative to the magnitude of both the true and predicted values. It is defined as
\begin{equation}
\mathrm{sMAPE}
=
\frac{100}{N}
\sum_{i=1}^{N}
\frac{
2 \left| y_i - \hat{y}_i \right|
}{
\left| y_i \right| + \left| \hat{y}_i \right| + \epsilon
},
\end{equation}
where $\epsilon$ is a small positive constant used to avoid division by zero. sMAPE is scale-independent and is commonly reported as a percentage. Lower sMAPE indicates better forecasting accuracy.

\begin{table}[ht]
\caption{Forecasting performance (sMAPE) across preprocessing pipelines and models. Best per model in bold. Columns in grey are IP-Models. Pipelines are composed of: \textbf{X} (Baseline/None), \textbf{V} (Boxcox transformation), \textbf{U} (First-order differencing), \textbf{S} (Seasonal differencing), and \textbf{Q} (Standard scaling).}
\centering
\small

\begin{tabularx}{\textwidth}{l G G G N N N N G G N G}

\toprule
Pipeline & ARIMA & ETS & THETA & LR & XGBst & MLP & NBTS & TNet & PTST & Informer & Chronos \\
\midrule
X    & 9.79 & 9.52 & \textbf{9.29} & 10.93 & 10.18 & 23.49 & 11.83 & 10.39 & 10.68 & 69.50 & 9.21 \\
V    & 12.43 & 9.21 & 9.38 & 10.73 & 9.88 & 12.15 & 10.92 & \textbf{9.71} & \textbf{9.98} & 81.52 & 9.43 \\
U    & 9.49 & 10.29 & 12.25 & 9.73 & 9.80 & 11.20 & 28.96 & 14.35 & 15.59 & 11.46 & 9.43 \\
S    & 9.78 & 13.09 & 13.58 & 10.95 & 9.85 & 12.18 & 15.68 & 13.77 & 17.16 & 13.27 & 10.40 \\
Q    & 9.54 & 9.26 & 9.44 & 10.92 & 12.82 & 14.64 & 15.93 & 10.07 & 10.38 & 15.95 & \textbf{9.21} \\
VU   & \textbf{9.28} & 9.84 & 11.15 & \textbf{9.67} & \textbf{9.32} & 10.48 & 28.47 & 13.19 & 13.41 & 10.84 & 9.22 \\
VS   & 9.58 & 12.25 & 12.59 & 10.71 & 9.50 & 10.04 & \textbf{10.20} & 12.16 & 12.36 & 10.61 & 10.24 \\
VQ   & 9.39 & \textbf{9.20} & 9.39 & 10.74 & 13.02 & 14.94 & 10.46 & 9.83 & 10.21 & 15.17 & 9.24 \\
US   & 10.33 & 15.90 & 18.93 & 13.44 & 12.85 & 14.76 & 35.38 & 24.16 & 24.71 & 18.54 & 14.57 \\
UQ   & 9.60 & 10.16 & 12.25 & 9.73 & 9.76 & \textbf{9.81} & 14.01 & 15.19 & 14.54 & 10.63 & 9.43 \\
SQ   & 9.93 & 12.98 & 13.57 & 10.95 & 9.85 & 10.38 & 10.26 & 13.44 & 13.73 & \textbf{10.30} & 10.40 \\
VUS  & 9.82 & 14.56 & 17.52 & 12.15 & 11.65 & 12.85 & 34.28 & 23.41 & 23.68 & 13.13 & 13.46 \\
VUQ  & 9.48 & 9.85 & 11.18 & 9.67 & 9.86 & 10.27 & 11.80 & 13.76 & 13.86 & 10.50 & 9.22 \\
VSQ  & 9.86 & 12.24 & 12.61 & 10.72 & 9.84 & 10.16 & 13.43 & 12.21 & 12.37 & 10.40 & 10.24 \\
USQ  & 11.98 & 15.69 & 18.91 & 13.44 & 13.52 & 14.29 & 16.11 & 25.13 & 24.52 & 14.32 & 14.57 \\
VUSQ & 11.33 & 14.56 & 17.52 & 12.15 & 12.32 & 12.55 & 33.64 & 24.38 & 23.76 & 13.22 & 13.46 \\
\bottomrule
\end{tabularx}
\label{tab:main_results_smape}
\end{table}

\section{Complementary Results}

Tables~\ref{tab:main_results_mse}, \ref{tab:main_results_rmse}, \ref{tab:main_results_mae}, and \ref{tab:main_results_smape} report forecasting performance when a single preprocessing pipeline is selected and applied uniformly to the entire dataset across MSE, RMSE, MAE, and sMAPE, respectively. This comprehensive evaluation mirrors the experimental setup described for Table~\ref{tab:main_results}, providing a broader perspective on model behavior under different loss functions and normalization schemes.

Across these metrics, the overall trends remain consistent with our primary findings: IP-Models demonstrate high robustness, typically achieving their peak performance under minimal preprocessing or simple scaling. In contrast, non-IP-Models exhibit more scattered performance, showing a general preference for differencing-based pipelines ($U$ or $S$). These patterns hold consistently across both scale-dependent (MSE, RMSE, MAE) and scale-independent (sMAPE) metrics, reinforcing that when architectures lack internal mechanisms to handle nonstationarity, external transformations are critical for optimal performance regardless of the evaluation metric used.

\subsection{Granular Preprocessing Preferences}
\label{appendix:heatmap}

\begin{figure}[ht]
\centering
\includegraphics[width=\textwidth]{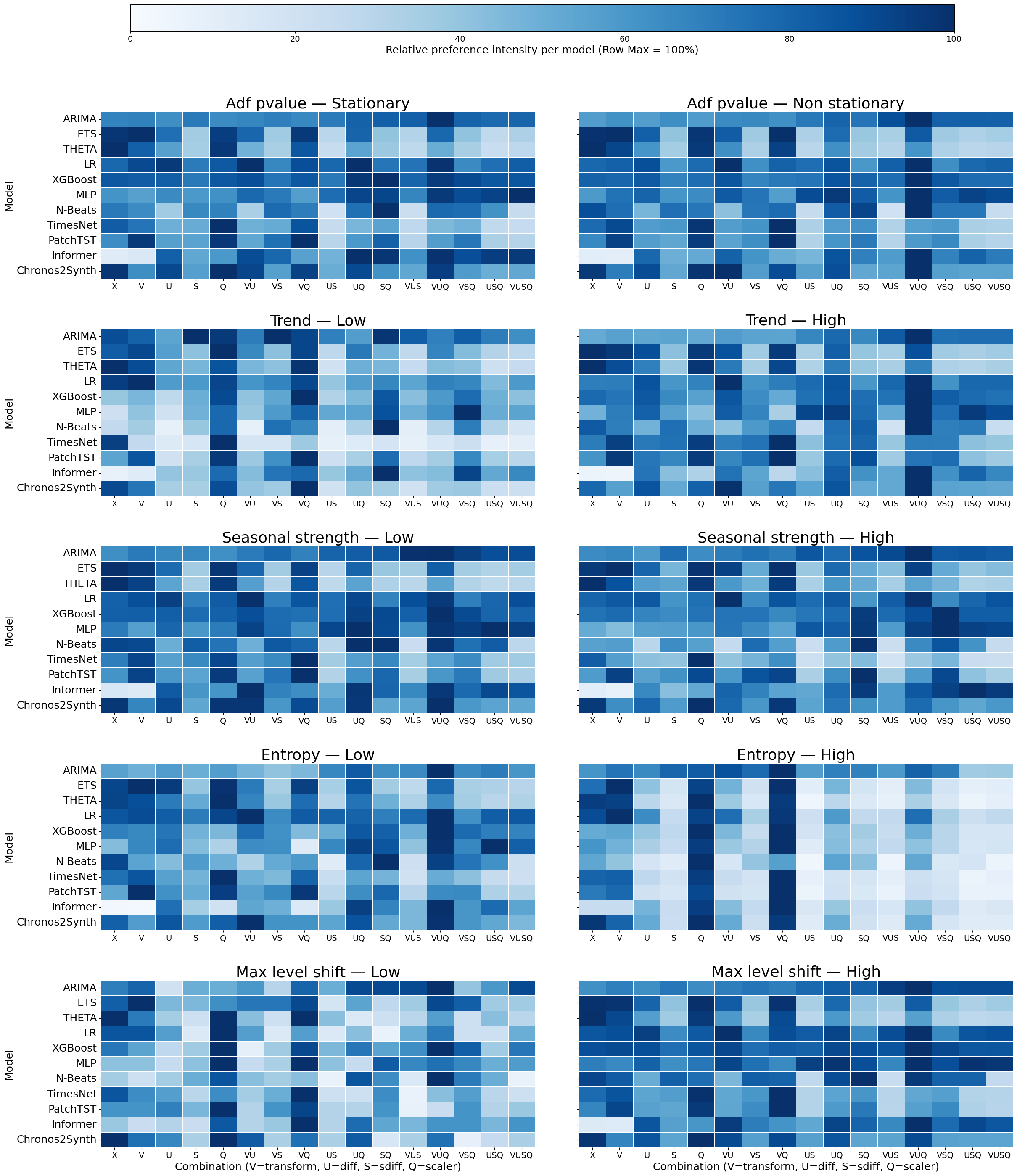}
\caption{Relative preference intensity for preprocessing pipelines achieving Top-3 MASE performance across time series subsets. Values are row-normalized per model to highlight the preferred pipelines for each architecture. Darker blue indicates a higher frequency of Top-3 performance.}
\label{fig:heatmap_test}
\end{figure}

Figure~\ref{fig:heatmap_test} provides a granular visualization of how often each preprocessing pipeline achieves Top-3 performance across different data characteristics. While Table~\ref{tab:recommendations} highlights the single absolute best pipeline, this heatmap complements those findings by revealing the broader distribution of highly competitive alternatives and the degree of preprocessing sensitivity for each architecture.

Several new patterns emerge from this expanded view. First, it highlights the difference between broad robustness and narrow dependence. Many IP-Models, such as ARIMA, THETA, and PatchTST, display broad horizontal bands of dark blue across multiple simple pipelines, such as $X$, $V$, $Q$, and $VQ$. This indicates that their performance remains highly competitive under various minimal transformations. Conversely, non-IP-Models like Informer, N-BEATS, and XGBoost often exhibit sharply concentrated preferences, seen as isolated dark blue blocks. For these architectures, selecting the correct pipeline is critical, as deviating from the optimal choice results in severe performance degradation.

Second, the heatmap visually captures the migration from simple to complex transformations. In the Low Trend and Stationary subsets, preference intensity is heavily concentrated on the left side of the grid (pipelines without differencing). In the High Trend and Non-Stationary subsets, non-IP-Models undergo a massive rightward shift, placing nearly all their preference weight on heavily differenced pipelines such as $VUQ$, $UQ$, and $USQ$.

Finally, the Entropy and Level Shift subplots provide additional nuance. Under High Entropy environments where series are noisier and more complex, models tend to favor simpler, less aggressive pipelines to avoid amplifying noise. Conversely, in Low Entropy scenarios with clearer structural predictability, models demonstrate a stronger capacity to leverage complex, multi-step pipelines successfully.

\subsection{No Pipeline x Oracle Deltas}
\label{appendix:nopipelinexoracle}
Table~\ref{tab:perc_gain} reinforces this point by quantifying the percentage gain from the "No Pipeline" to the "Oracle" configuration. Although the variance is high, reflecting the diversity of time series domains, a clear pattern is observable: models that lack internal preprocessing mechanisms exhibit the largest relative improvements. In contrast, IP-Models show comparatively smaller gains, consistent with their architectural design.

\begin{table}[ht]
\caption{MASE for each model across "No Pipeline" and "Oracle" configurations, and percentage gain. Positive gain means an increase in performance when using "Oracle". Grey rows indicate IP-Models.}
\centering
\small
\begin{tabular}{lccc}
\toprule
Model & MASE (No Pipeline) & MASE (Oracle) & Pct Gain (\%) \\
\midrule
\rowcolor{gray!12}
ARIMA     & 2.35 {\tiny $\pm$ 3.36} & 1.38 {\tiny $\pm$ 1.49} & 33.13 {\tiny $\pm$ 26.34} \\

\rowcolor{gray!12}
ETS       & 2.25 {\tiny $\pm$ 3.61} & 1.44 {\tiny $\pm$ 1.69} & 27.45 {\tiny $\pm$ 25.93} \\

\rowcolor{gray!12}
THETA     & 2.22 {\tiny $\pm$ 2.11} & 1.41 {\tiny $\pm$ 1.46} & 29.17 {\tiny $\pm$ 27.34} \\

LR        & 2.87 {\tiny $\pm$ 5.96} & 1.48 {\tiny $\pm$ 1.65} & 33.98 {\tiny $\pm$ 27.71} \\

XGBoost   & 2.75 {\tiny $\pm$ 3.77} & 1.24 {\tiny $\pm$ 1.24} & 41.48 {\tiny $\pm$ 28.68} \\

MLP       & 31.88 {\tiny $\pm$ 116.60} & 1.13 {\tiny $\pm$ 1.04} & 59.21 {\tiny $\pm$ 31.51} \\

N-BEATS   & 4.65 {\tiny $\pm$ 44.23} & 1.13 {\tiny $\pm$ 1.03} & 44.86 {\tiny $\pm$ 29.47} \\

\rowcolor{gray!12}
TimesNet  & 2.65 {\tiny $\pm$ 2.17} & 1.32 {\tiny $\pm$ 1.22} & 39.64 {\tiny $\pm$ 30.15} \\

\rowcolor{gray!12}
PatchTST  & 2.77 {\tiny $\pm$ 2.33} & 1.31 {\tiny $\pm$ 1.24} & 41.39 {\tiny $\pm$ 30.48} \\

Informer  & 106.75 {\tiny $\pm$ 500.84} & 1.14 {\tiny $\pm$ 0.93} & 86.73 {\tiny $\pm$ 21.54} \\

\rowcolor{gray!12}
Chronos2Synth & 2.29 {\tiny $\pm$ 2.86} & 1.39 {\tiny $\pm$ 1.40} & 31.39 {\tiny $\pm$ 25.68} \\
\bottomrule
\end{tabular}
\label{tab:perc_gain}
\end{table}

\subsection{Per-Model Ranking Analysis}
\label{appendix:ranking}

\begin{figure}[h]
\centering
\includegraphics[width=0.65\textwidth]{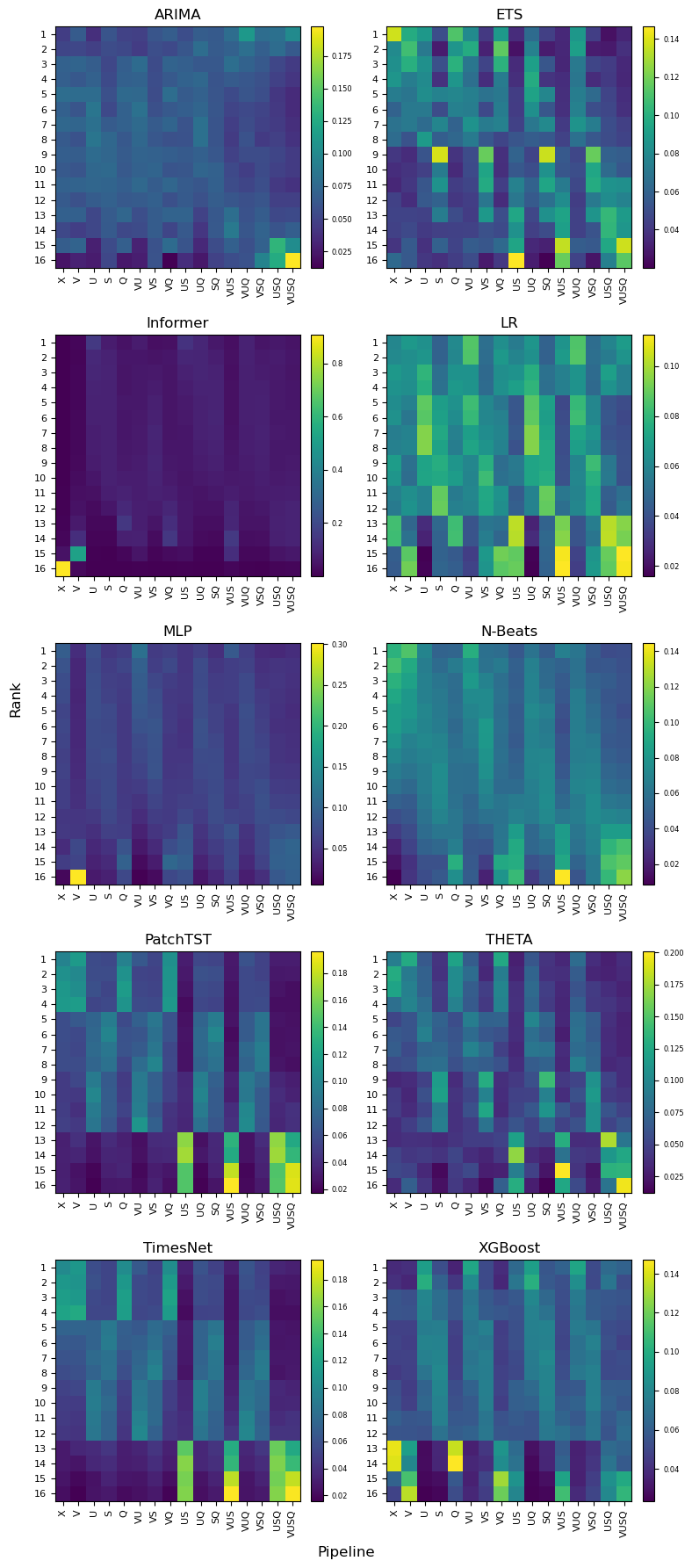}
\caption{Heatmaps illustrating the distribution of each preprocessing pipeline per Rank across all models.}
\label{fig:rank_heatmap2}
\end{figure}

Instead of average rank, Figure~\ref{fig:rank_heatmap2} shows the frequency that each pipeline achieves a certain rank. This figure gives a more detailed view of the patterns observed on the previous heatmap. IP-Models (except ARIMA) show a clear preference for $X$, $V$, $Q$, and $VQ$, where those 4 pipelines achieve Top-4 more often than any other pipeline. Non-IP-Models, again, have more varied preferences, without a single pattern in common. XGBoost, Informer, and MLP achieved higher ranks more often when using differencing such as $U$, $VU$, $UQ$, and $VUQ$. N-BEATS follows a similar pattern but also performs well without a preprocessing pipeline.

\subsection{Quantifying Cost-Related Trade-offs}
\label{appendix:cost}

As discussed previously, preprocessing does not significantly increase the computational cost per series. To thoroughly evaluate this, we compare the cost implications of models trained locally (one model per series) versus globally (one model across all series).

Table~\ref{tab:training_time_local} presents the training time differences between the No Pipeline (X) and Oracle configurations under the local training paradigm. For most models, the absolute difference remains marginal. 

\begin{table}[ht]
\caption{Training time comparison between Oracle and No-Pipeline configurations for local models. $\Delta$ is computed as No Pipeline (X) minus Oracle. Grey rows indicate IP-Models.}
\centering
\small
\setlength{\tabcolsep}{4pt}
\begin{tabular}{lccccc}
\toprule
Model & Oracle Mean (s) & X Mean (s) & $\Delta$ Mean (s) & $\Delta$ Std (s) & $\Delta$ (\%) \\
\midrule
\rowcolor{gray!12}
ARIMA    & 264.16 & 157.74 & -106.41  & 531.22 & -40.28 \\
\rowcolor{gray!12}
ETS      & 0.196  & 0.181  & -0.015   & 0.054  & -7.75  \\
\rowcolor{gray!12}
THETA    & 0.120  & 0.120  & -0.00003 & 0.0021 & -0.03  \\
LR       & 0.139  & 0.139  & -0.00009 & 0.0064 & -0.07  \\
XGBoost  & 3.393  & 3.115  & -0.277   & 1.161  & -8.17  \\
MLP      & 18.78  & 19.12  & 0.338    & 1.956  & 1.80   \\
N-BEATS  & 24.85  & 25.10  & 0.251    & 1.402  & 1.01   \\
\rowcolor{gray!12}
TimesNet & 175.32 & 177.45 & 2.134    & 11.11  & 1.22   \\
\rowcolor{gray!12}
PatchTST & 27.27  & 27.62  & 0.347    & 1.921  & 1.27   \\
Informer & 196.23 & 199.36 & 3.132    & 16.30  & 1.60   \\
\bottomrule
\end{tabular}
\label{tab:training_time_local}
\end{table}

\begin{figure}[h]
\centering
\includegraphics[width=0.65\textwidth]{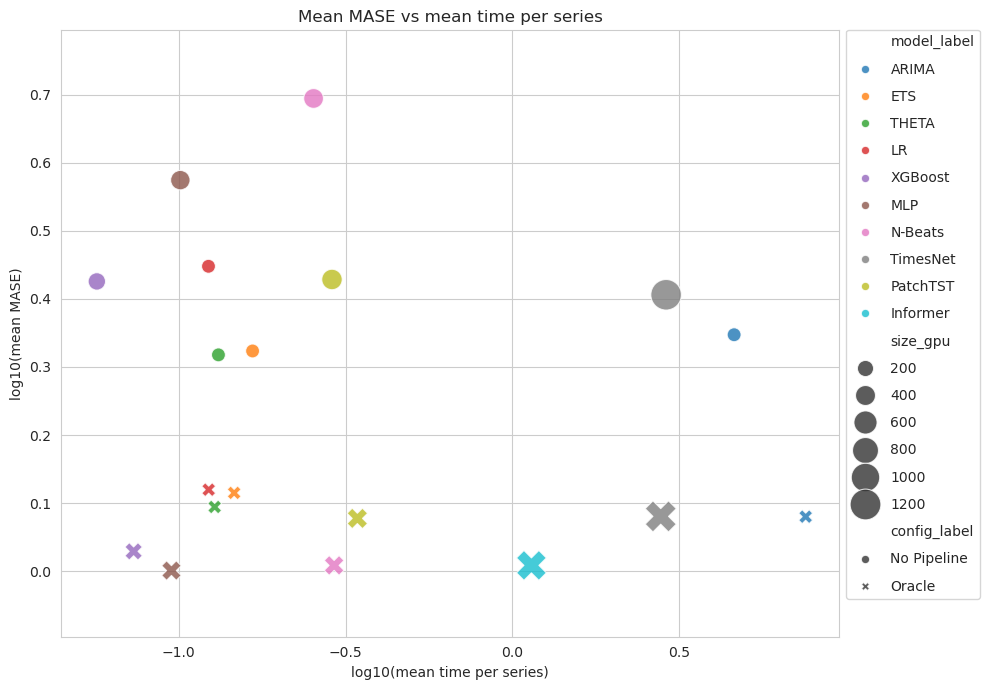}
\caption{Mean MASE vs. mean training time for global models. Marker size indicates peak GPU memory (baselined at 250MB for CPU-only models).}
\label{fig:mase_time_global}
\end{figure}

When scaling to global models, machine learning and deep learning architectures are trained jointly across the entire dataset. This global approach amortizes the training cost, allowing models like XGBoost and MLP to become remarkably fast per series. Figure~\ref{fig:mase_time_global} presents the performance versus cost trade-off for these globally trained models. In this global regime, deep learning and machine learning models efficiently leverage cross-series information without suffering the massive overhead of per-series fitting required by classical methods.

Table~\ref{tab:training_time} quantifies the training time differences in the global paradigm. Similar to the local results, the absolute time differences remain negligible for almost all models, reaffirming that applying preprocessing pipelines provides a performance boost with virtually no practical impact on training time.

\begin{table}[ht]
\caption{Training time comparison between Oracle and No-Pipeline configurations for global models. $\Delta$ is computed as No Pipeline (X) minus Oracle. Grey rows indicate IP-Models.}
\centering
\small
\setlength{\tabcolsep}{4pt}
\begin{tabular}{lccccc}
\toprule
Model & Oracle Mean (s) & X Mean (s) & $\Delta$ Mean (s) & $\Delta$ Std (s) & $\Delta$ (\%) \\
\midrule
\rowcolor{gray!12}
ARIMA    & 7.56   & 4.61   & -2.94    & 88.68  & -63.83 \\
\rowcolor{gray!12}
ETS      & 0.15   & 0.17   & 0.02     & 0.08   & 11.95  \\
\rowcolor{gray!12}
THETA    & 0.13   & 0.13   & 0.00     & 0.08   & 2.56   \\
LR       & 0.12   & 0.12   & -0.00    & 0.00   & -0.15  \\
XGBoost  & 0.07   & 0.06   & -0.02    & 0.01   & -28.89 \\
MLP      & 0.10   & 0.10   & 0.01     & 0.02   & 5.95   \\
N-BEATS  & 0.29   & 0.25   & -0.04    & 0.03   & -15.21 \\
\rowcolor{gray!12}
TimesNet & 2.78   & 2.89   & 0.11     & 1.83   & 3.65   \\
\rowcolor{gray!12}
PatchTST & 0.34   & 0.29   & -0.06    & 0.03   & -19.13 \\
Informer & 1.14   & 2.39   & 1.25     & 0.55   & 52.36  \\
\bottomrule
\end{tabular}
\label{tab:training_time}
\end{table}

\section{PATM}

\label{appendix:patm}

To support reproducibility and enable metalearning analyses, all experimental outputs are organized into a structured metadataset. As formalized in Equation~\ref{eq:patm}, each record corresponds to a single experimental tuple $(i, j, k)$, capturing the interaction between a time series, a preprocessing pipeline, and a forecasting model.

\subsection{File Organization}

The metadataset is distributed as a collection of CSV files, separated by dataset frequency. For each M4 subset, we provide:
(i) a \textit{PATM file}, containing experiment-level results, and
(ii) a \textit{feature file}, containing static descriptors of each time series.

The complete set of files generated in this study is:
\begin{itemize}
    \item \texttt{patm\_daily.csv}, \texttt{patm\_hourly.csv}, \texttt{patm\_quarterly.csv}, \texttt{patm\_weekly.csv}
    \item \texttt{m4\_daily\_features.csv}, \texttt{m4\_hourly\_features.csv}, \texttt{m4\_quarterly\_features.csv}, \texttt{m4\_weekly\_features.csv}
\end{itemize}

This separation avoids redundancy, as time series features are invariant across models and preprocessing pipelines, while PATM records are defined per experiment.

\subsection{PATM Schema}

Each row in a PATM file corresponds to a single forecasting experiment. Table~\ref{tab:patm_schema} summarizes the schema.

\begin{table}[h]
\centering
\small
\begin{tabular}{ll}
\toprule
\textbf{Column} & \textbf{Description} \\
\midrule
\texttt{series\_id} & Unique identifier of the time series \\
\texttt{model} & Forecasting model (e.g., ARIMA, MLP, PatchTST) \\
\texttt{pipeline} & Encoded preprocessing pipeline (e.g., X, VUQ) \\
\texttt{transformation} & Variance-stabilizing transformation (V) \\
\texttt{diff} & First-order differencing (U) indicator \\
\texttt{sdiff} & Seasonal differencing (S) indicator \\
\texttt{scaler} & Scaling method (Q) \\
\texttt{mape}, \texttt{mse}, \texttt{rmse}, \texttt{mae}, \texttt{smape}, \texttt{mase} & Forecasting performance metrics \\
\texttt{forecast} & Serialized forecast output (test horizon) \\
\texttt{time} & Training and inference time \\
\texttt{gpu\_peak\_memory\_mb} & Peak GPU memory usage (MB) \\
\texttt{seed} & Random seed \\
\texttt{pid} & Process identifier \\
\texttt{uuid} & Unique experiment identifier \\
\bottomrule
\end{tabular}
\caption{Schema of the Preprocessing-Aware Time Series Metadataset (PATM).}
\label{tab:patm_schema}
\end{table}

The schema captures both predictive performance and computational cost, enabling multi-objective evaluation.

\subsection{Example Record}

An example row from a PATM file is shown below:

\begin{verbatim}
series_id: 1
model: PatchTST
pipeline: none-diff-none-none
transformation: none
diff: diff
sdiff: none
scaler: standard_scaler
mape: 0.018365					
mse: 24334.813340
rmse: 155.996196
mae: 124.343401
smape: 1.819308
mase: 0.823671
forecast: [6852.52620975, 6859.60718623, 6883.1543111, ...]
time: 1.299543
gpu_peak_memory_mb: 370.0
seed: 42
pid: 807252
uuid: 0acce284-26da-43b9-8c33-68dde290e4d9
\end{verbatim}

This entry indicates that the PatchTST model was trained on series 1 using a pipeline composed of first-order differencing (U).

\subsection{Feature Files}

Each \texttt{m4\_<frequency>\_features.csv} file contains one row per time series, with the full set of descriptors described in Appendix~\ref{tab:features}. These features can be joined with PATM files via the \texttt{series\_id} field, enabling analyses of how time series characteristics influence preprocessing effectiveness and model performance.

\paragraph{Usage.}
The metadataset can be interpreted as a relational structure where PATM defines experiment-level observations and feature files define series-level attributes. This design supports meta-learning, preprocessing recommendation, and performance modeling tasks.

\end{document}